\documentclass[journal]{IEEEtran}
\usepackage{amsfonts}

\usepackage{url}
\usepackage{graphicx}
\usepackage{amsmath}
\usepackage{amssymb}
\usepackage{ifpdf}
\usepackage{cite}
\usepackage{array}
\usepackage[usenames,dvipsnames,svgnames,table]{xcolor}
\usepackage[numbers,sort,compress]{natbib}

\begin{document}

\title{Multi-Target Tracking with Time-Varying Clutter Rate and Detection
Profile: Application to Time-lapse Cell Microscopy Sequences}
\author{Seyed~Hamid~Rezatofighi,~\IEEEmembership{Student Member,~IEEE,}
Stephen~Gould,~\IEEEmembership{Member,~IEEE,} Ba~Tuong~Vo, Ba-Ngu~Vo, Katarina Mele and~Richard~Hartley,~%
\IEEEmembership{Fellow,~IEEE}\thanks{%
	Copyright (c) 2010 IEEE. Personal use of this material is permitted. However, permission to use this material for any other purposes must be obtained from the IEEE by sending a request to pubs-permissions@ieee.org.}\thanks{%
H. Rezatofighi is with the School of Computer Science, The University of Adelaide, Adelaide SA 5005, Australia (e-mail:
hamid.rezatofighi@adelaide.edu.au).}\thanks{%
S. Gould and R. Hartley are with the College of
Engineering and Computer Science, The Australian National University,
Canberra ACT 2601, Australia (e-mail: stephen.gould@anu.edu.au, richard.hartley@anu.edu.au).}\thanks{%
B.-N. Vo and B.-T. Vo are with Department of Electrical and Computer
Engineering, Curtin University of Technology, Perth WA 6845, Australia and are supported in part by the Australian Research Council under Discovery Project DP120102343 (e-mail: ba-ngu.vo@curtin.edu.au; ba-tuong.vo@curtin.edu.au).} \thanks{%
K. Mele is with Computational Informatics, CSIRO, North Ryde NSW 2113, Australia (e-mail:
katarina.mele@csiro.au).}}
\maketitle

\begin{abstract}
Quantitative analysis of the dynamics of tiny cellular and sub-cellular structures, known as particles, in time-lapse cell microscopy sequences requires the development of a reliable multi-target tracking method capable of tracking numerous similar targets in the presence of high levels of noise, high target density, complex motion patterns and intricate interactions.
In this paper, we propose a framework for tracking these structures based on the random finite set Bayesian filtering framework. We focus on challenging biological applications where image characteristics such as noise and background intensity change during the acquisition process. Under these conditions, detection methods usually fail to detect all particles and are often followed by missed detections and many spurious measurements with unknown and time-varying rates. To deal with this, we propose a bootstrap filter composed of an estimator and a tracker. The estimator adaptively estimates the required
meta parameters for the tracker such as clutter rate and the detection probability of the targets, while the tracker estimates the state of the targets. Our results show that the proposed approach can outperform state-of-the-art particle trackers on both synthetic and real data in this regime.
\end{abstract}

\markboth{IEEE TRANSACTIONS ON MEDICAL IMAGING, VOL. 34, NO. 6, JUNE 2015}{Shell \MakeLowercase{\textit{et al.}}: Bare Demo of IEEEtran.cls
for Journals}

\begin{IEEEkeywords}
Multi-target tracking, particle tracking, fluorescence microscopy, Bayesian estimation,
random finite set, CPHD, clutter rate, detection probability.
\end{IEEEkeywords}

\IEEEpeerreviewmaketitle

\section{Introduction}

\IEEEPARstart{T}{he} ability to accurately monitor cellular and sub-cellular structures in
their native biological environment has enormous potential in addressing
open questions in cell biology. In various applications, one of the key steps for understanding biological phenomena is to assess the motion of these structures.
Recent developments in time-lapse cell microscopy imaging systems
have had a great impact on the analysis of these dynamics. However, visual inspection of sequences acquired
by these imaging techniques requires manual tracking of many tiny structures in numerous noisy images. Thus,
automated multi-target tracking methods have been extensively used in different
biological applications in the last
decade~\cite{matov2011optimal,bonneau2005single,mashanov2007automatic,jaqaman2008robust,dewan2011tracking,sbalzarini2005feature,padfield2011coupled,meijering2006tracking,house2009tracking,liang2011expectation,nguyen2011tracking,genovesio2006multiple,yang2012new,chenouard2013multiple,juang2009tracking,rezatofighi2013multiple,wood2012simplified,feng2011multiple,rezatofighi2012application,li2008cell,smal2008multiple,smal2008particle,godinez2009deterministic,hoseinnezhad2012visual,yuan2012object}. 
\begin{figure}[t]
\begin{minipage}[b]{.49\linewidth}
  \centering
  \centerline{\includegraphics[height=4cm]{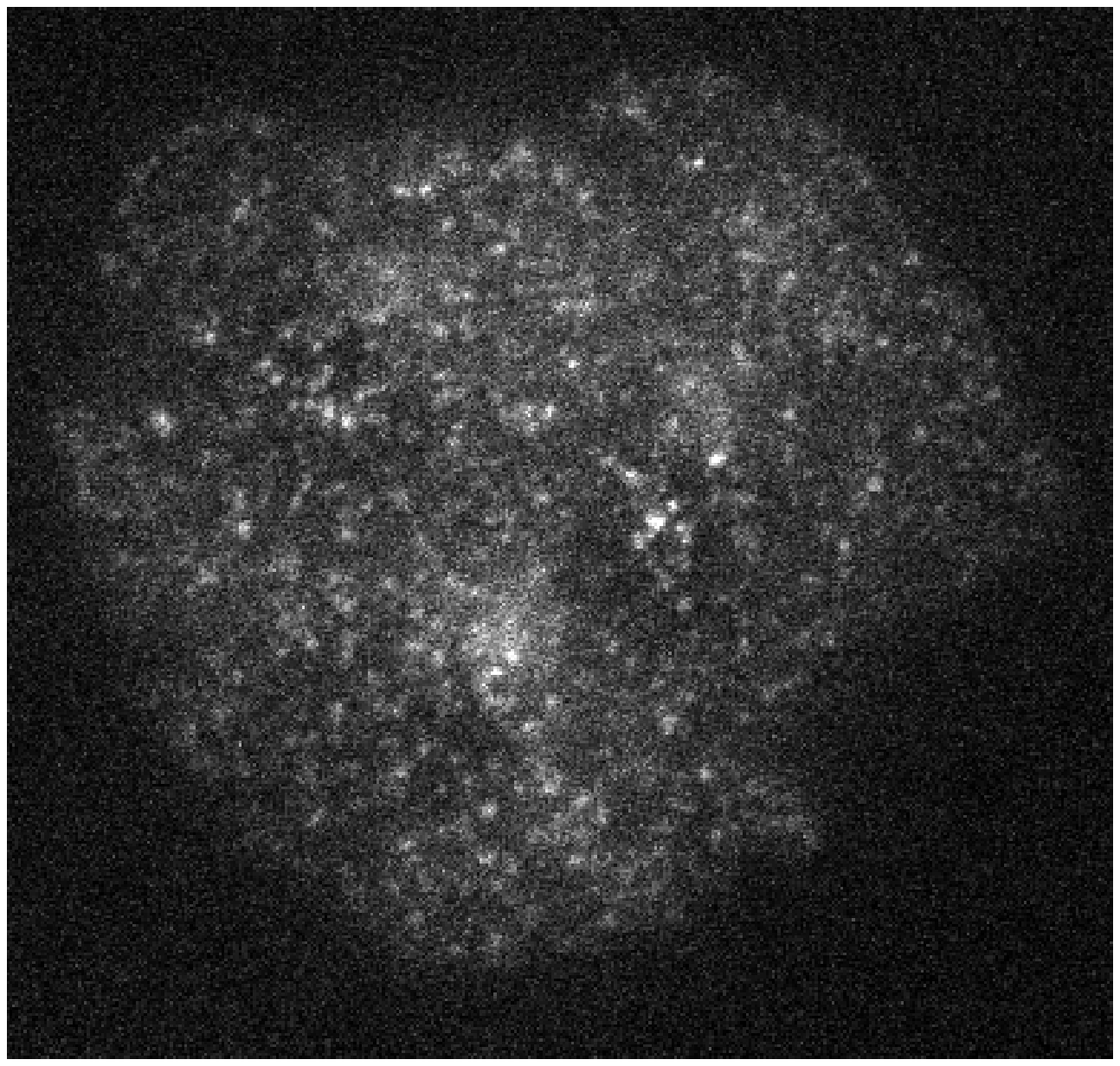}}
  \centerline{(a)}\medskip
\end{minipage}
\hfill
\begin{minipage}[b]{0.49\linewidth}
  \centering
  \centerline{\includegraphics[height=4cm]{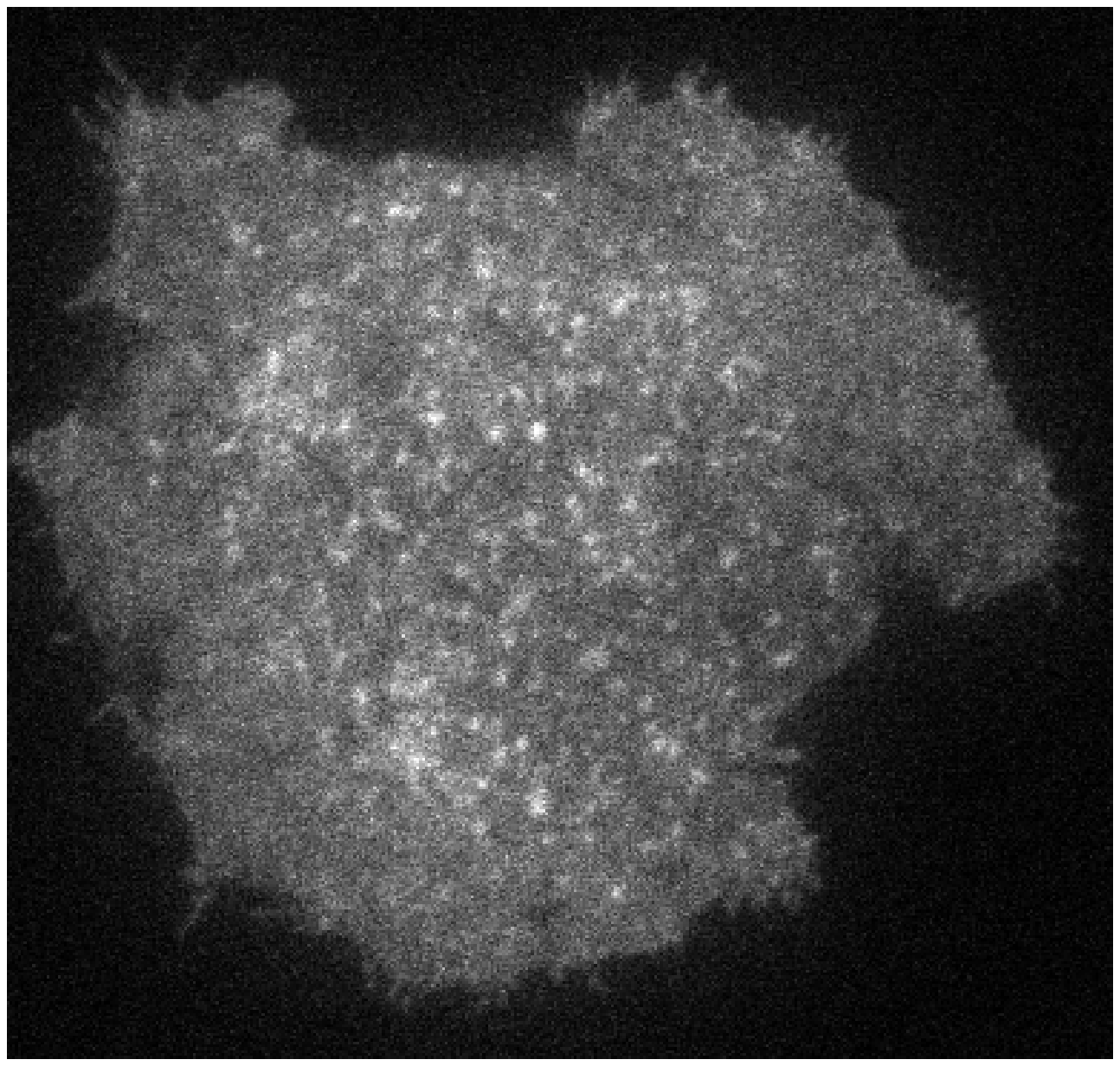}}
  \centerline{(b)}\medskip
\end{minipage}
\caption{ Two images of a TIRFM sequence visualizing fluorescently tagged
vesicles (bright spots) close to the plasma membrane of a pancreatic beta cell
(a) before and (b) after injection of insulin. Clearly, the background
intensity and noise level noticeably increase during the acquisition.}
\label{fig:TIRFM}
\end{figure}

Despite significant technical advances made in automatically
tracking moving objects, tracking microscopic structures, known as particles~\cite{chenouard2014objective}, remains a challenging
task due to the complex nature of biological sequences. The
microscopic sequences are usually populated with similar tiny
structures having intricate motion patterns and sophisticated
interactions with other structures. Moreover, the structures may enter or disappear from the field of view or be occluded by other objects. In addition in some imaging techniques, i.e. fluorescence
microscopy imaging, the sequences are contaminated with a high
degree of noise. Under these conditions, detection methods usually fail to detect all particles and generate many spurious measurements (clutter)~\cite{Smal2010quantitative,chenouard2013multiple}. To be successfully applied in many biological applications, multi-target tracking methods should be able to track an unknown and time-varying
number of similar particles in the presence of clutter and
detection uncertainty.

To this end, many particle tracking approaches have been proposed in literature~\cite{matov2011optimal,bonneau2005single,mashanov2007automatic,jaqaman2008robust,dewan2011tracking,sbalzarini2005feature,padfield2011coupled,meijering2006tracking,house2009tracking,liang2011expectation,nguyen2011tracking,genovesio2006multiple,yang2012new,chenouard2013multiple,juang2009tracking,rezatofighi2013multiple,wood2012simplified,feng2011multiple,rezatofighi2012application,li2008cell,smal2008multiple,smal2008particle,godinez2009deterministic,hoseinnezhad2012visual}. Some of the most popular particle tracking approaches are based on detection followed by a deterministic linking procedure. Here, each particle is separately detected in each frame. Then, a deterministic
solution, e.g. an optimization technique, links the corresponding
targets between frames~\cite{mashanov2007automatic,jaqaman2008robust,dewan2011tracking,sbalzarini2005feature,padfield2011coupled}. The performance of these algorithms
is often sensitive to the detection algorithm and may degrade in the
presence of complex target dynamics and highly cluttered detections resulting from very noisy sequences.

Bayesian filtering approaches are another class of tracking algorithms that have become
popular for particle tracking applications in recent years~\cite{liang2011expectation,nguyen2011tracking,genovesio2006multiple,yang2012new,chenouard2013multiple,juang2009tracking,hoseinnezhad2012visual,rezatofighi2013multiple,wood2012simplified,feng2011multiple,rezatofighi2012application,li2008cell,smal2008multiple,smal2008particle,godinez2009deterministic}. These approaches better deal with tracking
multiple particles in cluttered measurements by incorporating prior knowledge of target dynamics and
measurement models~\cite{chenouard2013multiple,juang2009tracking,rezatofighi2012application,wood2012simplified,rezatofighi2013multiple,feng2011multiple}. However, knowledge of the
clutter rate and detection profile of the chosen detection method, are of critical importance in
Bayesian multi-target tracking. 

Most existing multi-target tracking solutions assume known and
fixed detection and false alarm parameters. However, these parameters cannot be computed in many practical applications and worse, it is not even
known whether they are time-invariant. In biological imaging techniques, the noise characteristic and the background intensity of sequences
may change during the
acquisition process which make detection profile and clutter rate time-variant. For instance, injection of a
stimulus such as insulin into pancreatic beta cells increases the noise
level and overall intensity of the sequences acquired by total internal
reflection fluorescence microscopy (TIRFM)~\cite{Burchfield2010} (Fig.~\ref{fig:TIRFM}). Thus, the ability of multi-target trackers to accommodate unknown clutter and
detection parameters is crucial in time-lapse cell microscopy since
mismatches in these model parameters inevitably result in erroneous tracking
outputs.

The key contribution of this paper is to propose an effective solution to the problem of tracking multiple maneuvering particles in unknown and time-varying false alarm and detection rates. To the best of our knowledge, this practical problem has not been discussed in the
biological signal processing literature so far. To address this challenging problem, we use the recent generation of Bayesian filters based on random finite set theory. This framework provides an elegant mathematical formulation for multi-target systems and allows us to deal with the aforementioned complexities. Our proposed
approach is based on the Cardinalized Probability Hypothesis Density (CPHD)
filter for unknown clutter rate and detection profile known as the $\lambda $%
-$p_{D}$-CPHD filter \cite{mahler2011cphd}. Clutter rate and detection
probabilities are estimated by the $\lambda$-$p_D$-CPHD filter bootstrapped
onto a CPHD filter that outputs target estimates. This bootstrap idea was inspired by~\cite{beard2013multi} which requires known and uniform probability of detection. Our
proposed solution can accommodate unknown and non-uniform probability of
detection. To cope with maneuvering motion of sub-cellular structures~\cite{rezatofighi2012application,feng2011multiple,smal2008multiple,wood2012simplified,rezatofighi2013multiple,li2008cell}, we also propose the multiple model implementation of the aforementioned filters in this paper. We will show that
the proposed method is able to deal with tracking maneuvering structures in the presence of unknown and
time-varying clutter rate and detection profile.

\section{Background}\label{Sec:Background}
The Bayesian estimation paradigm deals with the problem of inferring the state of a target from a sequence of noisy measurements. The state vector contains all relevant dynamic information about the target while the measurements contain what can be directly observed from the sequences. The notion of Bayes optimality is fundamental to the Bayesian estimation
paradigm and is well-defined for single-target tracking~\cite{mahler2007statistical}.
While Bayes optimal tracking techniques such as Kalman and particle filters
are formulated for a single-target, they can be algorithmically extended to track a
time-varying number of targets by combining them with data association~\cite{bar1987tracking,blackman1986multiple,hue2002tracking}.
However, the notion of Bayes optimality does not carry over. Nonetheless this
approach to multi-target tracking has been used in a wide range of
applications, including particle tracking~\cite{yang2012new,smal2008multiple,smal2008particle,chenouard2013multiple,rezatofighi2012application,feng2011multiple,nguyen2011tracking,genovesio2006multiple,godinez2009deterministic,li2008cell}%
.

The random finite set (RFS) approach, introduced by Mahler generalizes the
notion of Bayes optimality to multi-target system using RFS theory~\cite{mahler2007statistical}. The key difference with other approaches is that
the collection of target states at any given time is treated as a set-valued
multi-target state. This representation provides an elegant formulation for multi-target system which avoids the need for data association. The RFS approach has generated substantial interest in
recent years with the development of the Probability Hypothesis Density
(PHD) filter~\cite{mahler2003multitarget}, Cardinalized PHD (CPHD) filter~\cite{mahler2007phd}, multi-Bernoulli filters~\cite{vo2009cardinality,vo2010joint}, labeled
RFS filter~\cite{vo2013labeled} and a host of many applications~\cite{mahler2014statistical} including cell and particle tracking~\cite{rezatofighi2013multiple,wood2012simplified,juang2009tracking,hoseinnezhad2012visual}.

\subsection{RFS for multi-target tracking}

In the Bayesian estimation paradigm, the state and measurement are treated
as realizations of random variables which are suitable for single target and single measurement system. However, in a multi-target system, the collection of states and measurements can be naturally represented as finite sets. 

Let $x_{k,1},...,x_{k,N(k)}$ and $z_{k,1},...,z_{k,M(k)}$ be the
states of all $N(k)$ targets and all $M(k)$ measurements at time
$k$, respectively. Over time, some of these targets may disappear, new targets
may appear, and the surviving targets evolve to new states.
Moreover, due to poor detector performance, only some targets are detected
at each time step and many measurements are spurious detections (clutter). Thus, we can conveniently represent the targets and measurements at each time slice with two finite sets as
\begin{equation}
\begin{aligned}
X_{k}& =\{x_{k,1},\ldots ,x_{k,N(k)}\}, \\
Z_{k}& =\{z_{k,1},\ldots ,z_{k,M(k)}\}.
\end{aligned}
\end{equation}
Consequently, the concept of a random finite set (RFS) is required to cast the
multi-target estimation problem in the Bayesian framework. Intuitively, an RFS is a finite-set-valued random variable. What distinguishes an RFS from a random vector is that the number of constituent variables is random and the
variables themselves are random, distinct and unordered. Mahler's Finite Set
Statistics (FISST) provides powerful yet practical mathematical tools for
dealing with RFSs~\cite{mahler2003multitarget,mahler2007statistical}, based on the
notion of integration and density that is consistent with point process
theory~\cite{vo2005sequential}. The centerpiece of the RFS approach is the so called
Bayes multi-target filter, a generalization of the single-target (optimal)
Bayes filter to accommodate multiple targets based on multi-target dynamic and measurement models.

\subsubsection{Multi-Target Dynamical Model}

Given a multi-target state $X_{k-1}$ at time $k-1$,\ each target $x_{k-1}$ $\in X_{k-1}$ either continues to exist at time $k$ with probability $
p_{S,k}(x_{k-1})$ and moves to a new state $x_{k}$ with probability density $
f_{k|k-1}(x_{k}|x_{k-1})$, or dies\ with probability $1-p_{S,k}
\left( x_{k-1}\right) $ and takes on the value $\emptyset $. Thus, given a
state $x_{k-1}\in X_{k-1}$ at time $k-1$, its behavior at time $k$ is
modeled by a Bernoulli RFS, $S_{k|k-1}$, which has FISST density\footnote{For more detail on RFS density and probability density please see~\cite{mahler2007statistical}} given by 
\begin{equation}
\begin{aligned}
& p_{k|k-1}(S_{k|k-1}|x_{k-1})=\\
&\left\{
\begin{array}{ll}
1-p_{S,k}(x_{k-1})& \textrm{if } S_{k|k-1}=\emptyset,\\
p_{S,k}(x_{k-1})f_{k|k-1}(x_{k}|x_{k-1})& \textrm{if } S_{k|k-1}=\{x_k\}. 
\end{array}\right.
\end{aligned}
\end{equation}
Assuming that individual targets move independently, each of the above Bernoulli RFSs
are mutually independent conditioned on $X_{k-1}$. Thus the survival or death of all existing targets
from time $k-1$ to time $k$ is modeled by the multi-Bernoulli RFS
\begin{equation}
T_{k|k-1}(X_{k-1})=\bigcup\limits_{x_{k-1}\in X_{k-1}}S_{k|k-1}(x_{k-1}).
\end{equation}

The appearance of new targets at time $k$ is modeled by an RFS of
spontaneous births $\Gamma _{k}$ which is usually specified as an i.i.d. cluster RFS with
intensity function $\gamma _{k}$ and cardinality distribution $\rho _{\Gamma
,k}$\footnote{An i.i.d. cluster RFS can be statistically represented by its intensity and cardinality distribution. The cardinality distribution, $\rho(n)$, provides a discrete probability distribution over the
number of the constituent random variables, $n$, while the intensity distribution $v(x)$ at the point $x$ in the state space
is interpreted as the instantaneous expected number of the variables that
exist at that point~\cite{mahler2007statistical}.}. Consequently, the RFS multi-target state $X_{k}$ at time $k$ is given
by the union
\begin{equation}
X_{k}=T_{k|k-1}(X_{k-1})\cup \Gamma _{k}.
\end{equation}
The likelihood of the multi-target state $X_k$, at time $k$, is described by the multi-target transition density $f_{k|k-1}(X_k|X_{k-1})$ which incorporates target dynamics, births and deaths~\cite{mahler2007statistical}.

\subsubsection{Multi-Target Measurement Model}

Given a multi-target state $X_{k}$ at time $k$, each target $x_{k}$ $\in
X_{k}$, at time $k$, is either detected with probability $p_{D,k}\left(
x_{k}\right) $ and generates an observation $z_{k}$ with likelihood $%
g_{k}(z_{k}|x_{k})$, or missed with probability $1-p_{D,k}\left( x_{k}\right) $ and generates the value $\emptyset $, \ i.e.
each target $x_{k}\in X_{k}$ generates a Bernoulli RFS, $D_k$, which has FISST density given by 
\begin{equation}
p_k(D_{k}|x_{k})=
\left\{
\begin{array}{ll}
1-p_{D,k}(x_{k})& \textrm{if }D_{k}=\emptyset,\\
p_{D,k}(x_{k})g_{k}(z_{k}|x_{k})& \textrm{if } D_{k}=\{z_k\}. 
\end{array}\right.
\end{equation}

The set of measurements generated by $X_k$, at time $k$, is modeled by the multi-Bernoulli RFS

\begin{equation}
\Theta _{k}(X_{k})=\bigcup\limits_{x_{k}\in X_{k}}D_{k}(x_{k}).
\end{equation}

In addition, the sensor receives a set of false/spurious measurements or
clutter, modeled by an RFS $K_{k}$, which is usually specified as an i.i.d.
cluster RFS with intensity function $\kappa _{k}$ and cardinality distribution $\rho _{K,k}$~\cite{mahler2007statistical}. Consequently,
at time $k$, the multi-target measurement $Z_{k}$ generated by the
multi-target state $X_{k}$ is formed by the union%
\begin{equation}
Z_{k}=\Theta _{k}(X_{k})\cup K_{k}.
\end{equation}
The likelihood of the multi-target measurement $Z_k$, given the multi-target state, at time $k$, is described by the multi-target measurement likelihood $g_k(Z_k|X_{k})$ which incorporates detection uncertainty and clutter.

\subsubsection{Bayes Multi-Target Filter}

The multi-target posterior density $p_k(X_k|Z_{1:k})$ contains all statistical information about the multi-target state at time $k$ and can be recursively computed by the aforementioned multi-target transition density
and measurement likelihood using the following prediction and update equations.
\begin{equation}\label{eq:bayes multi-target prediction}
\begin{aligned}
p_{k|k-1}&(X_k| Z_{1:k-1})=\\
&\int f_{k|k-1}(X_k| X_{k-1})p_{k-1}(X_{k-1}|Z_{1:k-1})\delta X_{k-1},
\end{aligned}
\end{equation}%
\begin{equation}\label{eq:bayes multi-target update}
p_k(X_k|Z_{1:k})\propto g_k(Z_k|X_{k})p_{k|k-1}(X_{k}|Z_{1:k-1}), 
\end{equation}%
where $\int (\cdot)\delta X$ is the set integral~\cite{mahler2007statistical}.

This filter generalizes the notion of Bayes optimality to multi-target system
using RFS theory. Further details can be found in~\cite{mahler2007statistical}.

\subsection{The PHD filters}

Although the above framework provides an elegant Bayesian formulation of the
multi-target filtering problem, it can be computationally expensive for applications with large number of targets~\cite{mahler2003multitarget}. In~\cite{mahler2003multitarget}, Mahler
proposed to propagate the probability hypothesis density (PHD), or
posterior intensity distribution, of the targets $v_k(x_k)$ which is the first
statistical moment of the probability density function
$p_k(X_k|Z_{1:k})$. 
The primary weakness of the PHD filter is
a loss of higher order cardinality information which causes erratic
estimates in the number of targets especially when the density of
targets is high~\cite{mahler2007phd,vo2007analytic,rezatofighi2013multiple}.

Mahler~\cite{mahler2007phd} subsequently proposed the Cardinalized
PHD (CPHD) filter which jointly propagates the posterior intensity function, $v_k$,
and cardinality distribution, $\rho_k$. The propagation of the cardinality distribution makes the filter more robust in the estimation of the number of targets~\cite{mahler2007phd}. 

As with all Bayesian filtering frameworks, this filter also requires some predefined
models such as the
single target Markov transition density $f_{k|k-1}(\cdot|\cdot)$ for
the dynamic model, measurement likelihood $g_{k}(\cdot|\cdot)$ and
probabilities of survival $p_{S,k}(\cdot)$ and detection
$p_{D,k}(\cdot)$. Moreover, the models for new born targets and false alarms (clutter) are described by the birth cardinality distribution $\rho_{\Gamma,k}$ and the birth intensity function $\gamma_k$, and the clutter cardinality distribution $\rho_{K,k}$ and the clutter intensity function  $\kappa_k$, respectively~\cite{mahler2007phd}. Obviously, this filter requires knowledge
of the detection probability and clutter distributions similar to other Bayesian filters.

In~\cite{mahler2011cphd}, it was shown that the CPHD filter can also be reformulated such that it
jointly estimates clutter distributions and detection probability while tracking. We
refer to this version of the filter as the $\lambda $-$p_D$-CPHD filter.
However, the $\lambda $-$p_D$-CPHD cannot naturally perform as well as the CPHD filter
when exact knowledge of detection probability and clutter density is available. 


\section{The Bootstrap CPHD Filter}
\begin{figure}[t]
\centering
\includegraphics[width=8cm]{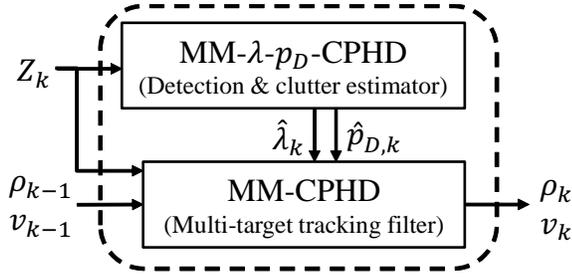}
\caption{A schematic of the proposed bootstrap filter.}
\label{fig:B-MM-CPHD}
\end{figure}

To propose an effective multi-target particle tracker for practical applications with unknown and time-varying clutter rate and detection profile, we borrow the
bootstrap idea proposed in~\cite{beard2013multi}. However, in addition to the clutter rate, we calculate the detection probability, using the $\lambda $-$p_D$-CPHD filter, and feed these to a robust multi-target tracking
filter such as the CPHD filter (Fig.~\ref{fig:B-MM-CPHD}).
We use the CPHD filter for target estimation because
it is a good trade-off between accuracy and speed. Nonetheless, it can be
replaced by any multi-target filter that requires knowledge of false alarm
and detection rate. In this paper,
we assume that $p_{D,k}$ and the clutter distributions, $\rho_{K,k}$ and $\kappa_k$, are the meta parameters to be estimated for the CPHD filter in
each time frame.  However, we will show that the estimation of the mean of the clutter rate, $\lambda_k$, is sufficient to uniquely define the $\rho_{K,k}$ and $\kappa_k$.

To deal with maneuvering dynamics, we assume that target motion
can be characterized by multiple dynamic models. Thus, here, we propose the multiple model (or jump-Markov) implementation of both the CPHD and
$\lambda$-$p_D$-CPHD filters and use these filters as the tracker and
the parameter estimator, respectively.

\subsection{Multiple Model CPHD filter}
Complex maneuvering motions can be often characterized by
multiple simpler dynamic models. The idea of tracking a target with multiple
motion models is to augment the state of the targets by the index of
the model~\cite{pasha2009gaussian}.

For notational convenience, we remove the time index $k$ in our
formulation throughout the paper. However, the random variables and distributions are
generally time-indexed. All random variables $(\cdot)_{k}$ at time
$k$ and $(\cdot)_{k-1}$ at time $k-1$ are simply denoted by
$(\cdot)$ and $\acute{(\cdot)}$, respectively. 

Let $\mathcal{X}$ and $\mathcal{R}=\{1,2,\cdots,R\}$
denote the kinematic state space and the multiple models state
space, respectively. Define the augmented state space as
$\underline{\mathcal{X}}=\mathcal{X}\times\mathcal{R}$, where
$\times$ denotes the Cartesian product. The underscore notation is
used for the augmented state space so that
$\underline{x}=[x,r]\in\underline{\mathcal{X}}$ denotes
$x\in\mathcal{X}$ for the kinematic state and $r\in\mathcal{R}$ for
the model state. The integral of a function
$\underline{f}:\underline{\mathcal{X}}\rightarrow \mathbb{R}$ is
given by
\begin{align}\label{eq:The integral of a function}
  \int_{\underline{\mathcal{X}}}\underline{f}(\underline{x})d\underline{x}&=\sum_{r=1}^{R}\int_{\mathcal{X}}\underline{f}(x,r)dx.
\end{align}
The measurement likelihood and probabilities of survival and
detection for the augmented state vector can be respectively written as
\begin{equation}\label{eq:Augmented state likelihood}
\underline{g}(z|\underline{x})= \underline{g}(z|x,r),
\end{equation}
\begin{equation}\label{eq:Augmented state pS}
\underline{p}_{S}(\underline{x})=\underline{p}_{S}(x,r),
\end{equation}
\begin{equation}\label{eq:Augmented state pD}
\underline{p}_{D}(\underline{x})=\underline{p}_{D}(x,r).
\end{equation}
Also, for the augmented transition
density and birth intensity, we have the following factored forms,
\begin{equation}\label{eq:Augmented state f}
\underline{f}(\underline{x}|\acute{\underline{x}})=\underline{f}(x,r|\acute{x},\acute{r})=f(x|\acute{x},r)\tau(r|\acute{r}),
\end{equation}
\begin{equation}\label{eq:Augmented state Gamma}
\underline{\gamma}(\underline{x})=\underline{\gamma}(x,r)=\gamma(x)\pi(r),
\end{equation}
where $\tau(r|\acute{r})$ is the probability transition from model
$\acute{r}$ to $r$ and $\pi(r)$ is the probability of birth for model $r$~\cite{pasha2009gaussian}. As described in Section~\ref{Sec:Background}, the filter also requires the knowledge of the birth cardinality distribution $\rho_{\Gamma}(n)$.

Clutter is usually assumed to be Poisson RFS and independent from the target state~\cite{mahler2007statistical,blackman1986multiple,bar1987tracking,hue2002tracking} with
\begin{equation}\label{eq:Clutter cardinality}
\rho_K(n)=Pois(n;\lambda),
\end{equation}
\begin{equation}\label{eq:Clutter intensity}
\kappa(z)=\lambda\mathcal{K}(z),
\end{equation}
where $\mathcal{K}(z)$ is a known probability distribution, e.g. uniform distribution, representing how the false alarms are distributed over the measurement space and $\int\mathcal{K}(z)dz=1$.
Therefore, given $\mathcal{K}(z)$, knowledge of the parameter $\lambda$
is sufficient to define the clutter distribution.

Substituting the aforementioned augmented model terms (Eqs.~\ref{eq:Augmented state likelihood}-\ref{eq:Clutter intensity}) into the
conventional CPHD equations~\cite{vo2007analytic} and using
Eq.~\ref{eq:The integral of a function} yields the recursive equations for
the multiple model or jump Markov CPHD filter (see Appendix~\ref{appendix:CPHD}).

Fig.~\ref{fig:MM-CPHD} depicts a simple graphical representation
of the proposed multiple model CPHD filter.
At each time step $k$,
the intensity $v_k$ and cardinality $\rho_k$ distributions are
predicted and updated from the intensity $v_{k-1}$ and cardinality
$\rho_{k-1}$ distributions in the previous time step $k-1$ and using
the augmented model terms (Eqs.~\ref{eq:Augmented state likelihood}-\ref{eq:Clutter intensity}). The intensity distribution provides
information on the state of the targets while the number of targets
can be estimated using the cardinality distribution. Specifically,
the number of targets is estimated using the posterior mode
$N_k=\operatorname{arg\,max} \rho_k(n)$~\cite{vo2007analytic}. Note
that this filter requires the clutter rate $\lambda$ and
the detection probability $p_{D}$ as prior knowledge.
\begin{figure}[t]
\centering
\includegraphics[width=7.5cm]{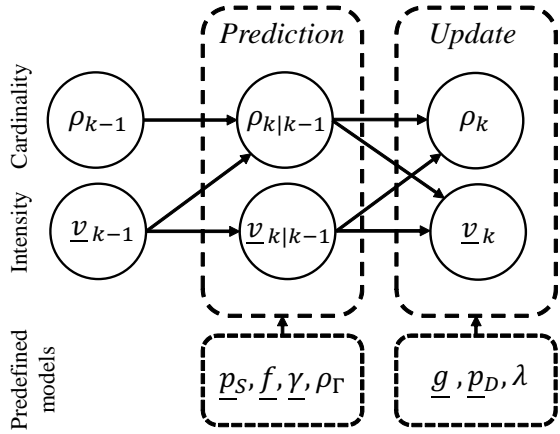}
\caption{A schematic of the proposed MM-CPHD filter.}
\label{fig:MM-CPHD}
\end{figure}
\subsection{Multiple Model $\lambda$-$p_D$-CPHD filter}

We borrow a key idea from Mahler \emph{et al.}~\cite{mahler2011cphd}
to develop a multiple model CPHD filter with unknown clutter rate
and detection profile. To estimate clutter, the idea is to model it as a random finite set of false targets which is statistically independent from the set of
actual targets. These false targets are defined by their own models
such as birth, death, survival and detection probabilities and
transition model. Therefore, the multi-target state is composed of a
disjoint combination of these two finite sets including actual
targets and clutter. This hybrid multi-target state allows us to track both actual and false targets simultaneously. To deal with multiple model dynamics and
unknown detection probability, the state of targets are augmented by
the index of the models and the unknown detection probability. Finally
this hybrid and augmented state is estimated from the sequence of finite
sets of measurements generated by both real targets and clutter
while accommodating multiple model dynamics and estimating the
detection probability and clutter rate.

Let $\mathcal{X}^{(1)}$ denote the state space for actual targets,
$\mathcal{X}^{(0)}$ denote the state space for clutter generators
and $\mathcal{X}^{(\Delta)}=[0,1]$, $\mathcal{R}=\{1,2,\cdots,R\}$
and $\mathcal{Q}=\{1,2,\cdots,Q\}$ denote the state space for the
unknown detection probability, and the multiple models for actual
targets and clutter, respectively. Define the hybrid and augmented
state space as
\begin{align}
  \underline{\ddot{\mathcal{X}}}
  &=\left(\mathcal{X}^{(1)}\times\mathcal{X}^{(\Delta)}\times\mathcal{R}\right)\uplus\left(\mathcal{X}^{(0)}\times\mathcal{X}^{(\Delta)}\times\mathcal{Q}\right)
\end{align}
where $\uplus$ denotes a
disjoint union. The double dot notation is used throughout to denote
a function or variable defined on the hybrid state space and the
underscore notation is used for the augmented state space.
Therefore, $\underline{\ddot{x}}\in\underline{\ddot{\mathcal{X}}}$
represents a hybrid and augmented state such that
$\underline{x}=(x,a,r)\in\underline{\mathcal{X}}^{(1)}=\mathcal{X}^{(1)}\times\mathcal{X}^{(\Delta)}\times\mathcal{R}$
and
$\underline{c}=(c,b,q)\in\underline{\mathcal{X}}^{(0)}=\mathcal{X}^{(0)}\times\mathcal{X}^{(\Delta)}\times\mathcal{Q}$
denotes the actual and clutter states comprising the kinematic state
and augmented components, respectively. The integral of a function
$\underline{\ddot{f}}:\underline{\mathcal{\ddot{X}}}\rightarrow
\mathbb{R}$ is given by
\begin{multline}\label{eq:The integral of hybrid function}
\int_{\underline{\ddot{\mathcal{X}}}}\underline{\ddot{f}}(\underline{\ddot{x}})d\underline{\ddot{x}}=\sum_{r=1}^{R}\int_{\mathcal{X}^{(\Delta)}}\int_{\mathcal{X}^{(1)}}\underline{\ddot{f}}(x,a,r)dxda+\\
\sum_{q=1}^{Q}\int_{\mathcal{X}^{(\Delta)}}\int_{\mathcal{X}^{(0)}}\underline{\ddot{f}}(c,b,q)dcdb.
\end{multline}

In theory, there may exist multiple model clutter generators.
However, in most tracking applications, it is assumed that the
clutter is uniformly distributed with Poisson
cardinality~\cite{mahler2007statistical,blackman1986multiple,bar1987tracking,hue2002tracking}.
As a result, multiple model clutter generators with uniform-Poisson
distribution can be substituted by a single model clutter generator
with uniform distribution and higher Poisson mean. Therefore, we
continue our derivations with the single model clutter generators by eliminating the variable $q$ from the equations. Deriving the equations for applications with multiple clutter generators and non-uniform distribution is straightforward.

In addition, since clutter generators are
identical and the false targets do not follow any motion pattern, it is reasonable to ignore any functional
dependence on the state of a clutter generator $c$,~\cite{mahler2011cphd}. As a result, the probabilities of survival and
detection and the measurement likelihood for the hybrid and augmented state vector can be respectively written as
\begin{align}\label{eq: hybrid Ps}
 \underline{\ddot{p}}_{S}(\underline{\ddot{x}})=\left\{
\begin{array}{ll}
p^{(1)}_{S}(x,r),&\underline{\ddot{x}}\in\underline{\mathcal{X}}^{(1)}\\
p^{(0)}_{S},& \underline{\ddot{x}}\in\underline{\mathcal{X}}^{(0)}
\end{array} \right.
\end{align}
\begin{align}
 \underline{\ddot{p}}_{D}(\underline{\ddot{x}})=\left\{
 \begin{array}{ll}
 a,&
\underline{\ddot{x}}\in\underline{\mathcal{X}}^{(1)}\\
b,&
\underline{\ddot{x}}\in\underline{\mathcal{X}}^{(0)}\\
 \end{array} \right.
\end{align}
\begin{align}
 \underline{\ddot{g}}(z|\underline{\ddot{x}})=\left\{
 \begin{array}{ll}
g(z|x,r),&
\underline{\ddot{x}}\in\underline{\mathcal{X}}^{(1)}\\
\mathcal{K}(z),&
\underline{\ddot{x}}\in\underline{\mathcal{X}}^{(0)}\\
 \end{array} \right.
\end{align}
where $\mathcal{K}(\cdot)$ is clutter density which is often assumed to be uniform distribution in the measurement space. As the survival probabilities and the measurement likelihoods are independent from the detection probabilities, the terms $a$ and $b$ are removed from the equations. Obviously, the detection probabilities are only dependent on $a$ and $b$. 

For the hybrid and augmented birth intensity and transition
density, we have the following factored forms,

\begin{align}
 \underline{\ddot{\gamma}}(\underline{\ddot{x}})=\left\{
 \begin{array}{ll}
\underline{\gamma}^{(1)}(x,a)\pi(r),&
\underline{\ddot{x}}\in\underline{\mathcal{X}}^{(1)}\\
\underline{\gamma}^{(0)}(b),&
\underline{\ddot{x}}\in\underline{\mathcal{X}}^{(0)}\\
 \end{array} \right.
\end{align}

\begin{align}\label{eq: hybrid f}
 \begin{array}{ll}
 \underline{\ddot{f}}(\underline{\ddot{x}}|\acute{\underline{\ddot{x}}})=\\
 \left\{\!\!\!\
 \begin{array}{ll}
f^{(1)}(x|\acute{x},r)f^{(\Delta)}(a|\acute{a},\acute{r})\tau(r|\acute{r}),&
\underline{\ddot{x}},\underline{\acute{\ddot{x}}}\in\underline{\mathcal{X}}^{(1)}\\
f^{(0)}(\underline{c}|\underline{\acute{c}}),&
 \underline{\ddot{x}},\underline{\acute{\ddot{x}}}\in\underline{\mathcal{X}}^{(0)}\\
 0,&\textrm{otherwise,}
 \end{array} \right.
\end{array}
\end{align}
where $f^{(\Delta)}(a|\acute{a},\acute{r})$ is the transition
density for the detection probability $a$ and 
$f^{(0)}(\underline{c}|\underline{\acute{c}})=f^{(\Delta)}(b|\acute{b})$ due to the
independence of false alarms from the clutter state $c$. Furthermore, the cardinality
distribution of birth for the hybrid space is
$\ddot{\rho}_{\Gamma}=\rho^{(1)}_{\Gamma}\ast\rho^{(0)}_{\Gamma}$~\cite{mahler2011cphd}.

By substituting the hybrid and augmented state space model terms
(Eqs.~\ref{eq: hybrid Ps}--\ref{eq: hybrid f}) into the conventional
CPHD equations~\cite{vo2007analytic} and using Eq.~\ref{eq:The
integral of hybrid function}, the recursive equations for this
filter can be calculated (see Appendix~\ref{appendix:lambda-CPHD}).

Fig.~\ref{fig:lambda-CPHD} shows a simple graphical
representation of the proposed multiple model $\lambda$-$p_D$-CPHD
filter.
At each time step $k$, the target's intensity
$\underline{v}^{(1)}_k$, the clutter intensity
$\underline{v}^{(0)}_k$ and the hybrid cardinality distribution
$\ddot{\rho}_k$ are predicted and updated from the target's
intensity $\underline{v}^{(1)}_{k-1}$, the clutter intensity
$\underline{v}^{(0)}_{k-1}$ and the hybrid cardinality distribution
$\ddot{\rho}_{k-1}$ in the previous time step $k-1$ and using the
hybrid model terms (Eqs.~\ref{eq: hybrid Ps}--\ref{eq: hybrid f}).
\begin{figure}[t]
\centering
\includegraphics[width=7.5cm]{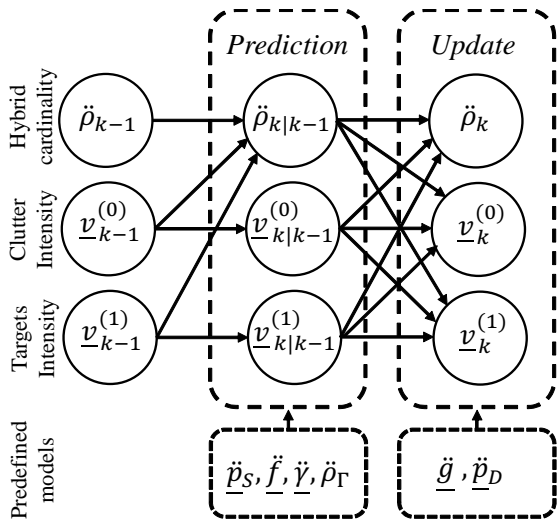}
\caption{A schematic of the proposed MM-$\lambda$-$p_D$-CPHD filter.}
\label{fig:lambda-CPHD}
\end{figure}

The posterior cardinality $\ddot{\rho}_k$ provides information on
the total number of real and clutter targets. Therefore, the number
of actual targets cannot be estimated using the posterior mode
$\ddot{N}_k=\operatorname{arg\,max} \ddot{\rho}_k(\ddot{n})$, since
$\ddot{\rho}_k(\ddot{n})$ includes both real targets and clutter. In
this filter, the posterior mean
$N^{(1)}_k=\sum_r\int\int\underline{v}^{(1)}_k(x,a,r)dxda$ is used for
estimation of the number of real targets~\cite{mahler2011cphd}.
Therefore, in contrast with the CPHD filter, this filter cannot benefit from propagation of the cardinality distribution for the
estimated number of the targets. Similar to the PHD filter, this
leads the erratic estimates of the number of targets which worsens
the tracking result. To this end, we use this filter only as the
estimator in our bootstrap filter. The estimated detection and mean
clutter are respectively calculated as follows~\cite{mahler2011cphd}.
\begin{equation}
\begin{aligned}
\hat{p}_{D,k}=&\langle\underline{v}_k^{(1)},\underline{p}_{D,k}^{(1)}\rangle, \\
\hat{\lambda}_k=&\langle\underline{v}_k^{(0)},\underline{p}_{D,k}^{(0)}\rangle,
\end{aligned}
\end{equation}
where $\langle \cdot\!,\!\cdot\rangle$ is the inner
product operator, $\underline{p}^{(1)}_{D,k}\!=\!a$ and
$\underline{p}^{(0)}_{D,k}\!=\!b$.

\section{Experimental results}

To evaluate the performance of our proposed filters, we apply them
to multi-target tracking in 2-D Total Internal Reflection
Fluorescence Microscopy (TIRFM) sequences. TIRFM is an imaging
technique that enables visualization of sub-cellular structures such
as vesicles that are on or close to the plasma membrane of
cells~\cite{Burchfield2010}. The vesicles are very tiny sub-cellular
structures and are seen in TIRFM sequences as small particles (bright spots)
moving with varying dynamics. These small particles appear and disappear from
the field of view and can be occluded by other structures. Due to
limitations in the TIRFM acquisition process, the sequences are
contaminated with a high level of noise. We consider the case where
the main characteristics of the sequences such as noise level and
the background intensity gradually increase over time
(Fig.~\ref{fig:TIRFM}). 

In this specific application, we are interested in the overall motion of the vesicles before and after injection of insulin, not motion of a single object. In this case, the primary concern is not how well a tracking approach maintains the identity of a tracked object over time. Instead, we are mainly concerned with how well a tracker avoids false and missed tracks.

We compare the results of our bootstrap filter against those of several popular state-of-the-art particle tracking methods including two reliable deterministic linking techniques, P-Tracker~\cite{sbalzarini2005feature} and U-Tracker~\cite{jaqaman2008robust}, and two robust detection based traditional Bayesian filtering approaches, MHT~\cite{chenouard2013multiple} and IMM-JPDA~\cite{rezatofighi2012application}.\footnote{The performance of the first three trackers, P-Tracker, U-Tracker and MHT, has been evaluated in a recent particle tracking challenge and their results have been reported in~\cite{chenouard2014objective}. According to this article, they can be assumed as three top performing methods for similar applications.}    
Our results are also compared against the result of a multiple model PHD
filter~\cite{rezatofighi2013multiple}. Finally, the results of our other derived filters such as the multiple-model CPHD (MM-CPHD) and $\lambda$-$p_D$-CPHD (MM-$\lambda$-$p_D$-CPHD) filters are also reported here in order to show the efficiency of our bootstrap idea.

\subsection{Setup and Implementation Details}

To fairly evaluate the performance of all tracking algorithms, the same detection lists were provided for all competing tracking methods. We chose the detector proposed in~\cite{rezatofighi2012new} as it performs reliably in our synthetic and real sequences.

Having ground truth, we can accurately calculate the clutter (false positive) and the detection probability (true positive rate) of the chosen detector for each time frame. True positive and false positive were calculated based on the optimal sub-pattern assignment between the ground truth and detected point sets. In this case, if the distance between two optimally assigned points is less than a predefined distance ($2$ pixel), the detected point is counted as true positive, otherwise it is a false positive.

For the MHT, IMM-JPDA,
MM-PHD and MM-CPHD filters, the average
number of false detections per frame $\bar{\lambda}$ and the mean value of the detection probability $\bar{p}_D$ (the optimal value for these parameters) were used as the predefined clutter rate and detection probability. However, accurate knowledge of these values is not possible in many practical applications and thus, the reported results for these filters are optimistic. 
For the MM-$\lambda$-$p_D$-CPHD
filter and similarly for our bootstrap (B-MM-CPHD) filter, the
clutter rate and detection probability are adaptively estimated using our proposed framework.

To ensure the validity of our experiments, for the independently implemented tracking methods such as MHT, P-Tracker and U-Tracker, we attempted to either estimate
their parameters from the ground truth or to find the values that resulted in their best performance. Moreover, the multiple motion model implementation of all competing tracking methods was used. For the IMM-JPDA and different PHD and CPHD filters, we chose identical state and measurement vectors
and dynamic models. We modeled the state of each
particle by its position, $x_x,x_y$, and velocity,
$\dot{x}_x,\dot{x}_y$. The measurements vector also contained the estimated position of the particles as $z=(\hat{x}_x,\hat{x}_y)$. To model maneuvering motion of
particles, two linear dynamics including random walk and small
acceleration motion models were
used~\cite{rezatofighi2012application,feng2011multiple}.

Since the target dynamics and
measurement models in this application can be properly expressed by multiple linear and Gaussian
terms, we used the Gaussian
mixture implementation~\cite{vo2007analytic} and the Beta-Gaussian mixture
approach~\cite{mahler2011cphd} for analytical implementations of the
MM-CPHD and 
MM-$\lambda$-$p_D$-CPHD filters, respectively. The birth
intensity distribution $\gamma(\cdot)$ for all the PHD and CPHD filters was set as a Gaussian
distribution centered at the image with a very high standard
deviation~\cite{rezatofighi2013multiple}. 

As previously discussed, the PHD and CPHD filters propagate the intensity distribution of all targets $v_k(x)$ in each time
frame $k$. By calculating the estimated number of targets $N_k$, the state of all targets in each time frame can be easily extracted using
$v_k(x)$~\cite{vo2006gaussian,vo2005sequential,vo2007analytic}\footnote{For example, in the Gaussian
mixture implementations, the mean of $N_k$ Gaussian terms with the highest weights in $v_k(x)$ can be used as the estimation of the state vector at each time frame~\cite{vo2006gaussian}.}. However, in this filtering framework, the temporal correspondences between the estimated
states is not considered. In other words, the output of the PHD and CPHD filters is a set of the
estimated states for each time frame without considering the identity of trajectories.
Thus, the dynamics of an individual target cannot be evaluated. To deal with this, some authors combine the PHD filter with a track management technique to maintain the identity of tracks~\cite{lin2006track,panta2007novel}. To avoid any computational burden due to a track management step, we simply use our tag propagation scheme~\cite{rezatofighi2013multiple}, which only propagates the identity of the intensity distributions for all the PHD and CPHD
filters. As it only considers the previous time step to propagate the identities~\cite{rezatofighi2013multiple}, this is not the most reliable approach for identity-to-track assignment. However, since we are interested in the overall, not individual, motion of the vesicles, we used this approach in this application.

\subsection{Evaluation Metric}
To qualitatively assess the performance of these tracking methods,
we used a recent and popular metric based on optimal
sub-pattern assignment (OSPA)~\cite{schuhmacher2008consistent} followed by an extension of this metric for multi-target tracking applications, known as OSPA-T~\cite{ristic2011metric}.   

\subsubsection{OSPA Metric}
This
	metric measures the distance between two sets of points, which can be
	the set of estimated tracks and the ground truth tracks at each time frame, and
	represents different aspects of multi-target tracking performance
	such as track accuracy, track truncation and missed or false tracks
	by a single value.

For two arbitrary sets $X=\{x_{1},...,x_{m}\}$ and
$Y=\{y_{1},...,y_{n}\}$, where $m\leq n$, the metric of order $p$ is
defined as
\begin{equation}\label{eq:OSPA}
D^c_p(X,\!Y)\!=\!\!\Biggl(\!\frac{1}{n}\Bigl(\min_{\theta\in\Theta_n}\!\!\sum_{i=1}^m\!\!
d^c_p(x_{i},y_{\theta_{i}})^p\!+
c^p(n-m)\Bigr)\!\Biggr)^{\!\!\frac{1}{p}},
\end{equation}
where $\Theta_n$ is all feasible sets of permutations on $Y$ and
\begin{equation}\label{eq:location OSPA}
d^c_p(x,y)=\min(c,\|x-y\|_p),
\end{equation}
where $c>0$ is the cut-off parameter.
For $m>n$, $D^c_p(Y,X)$ is calculated, instead.

In Eq.~\ref{eq:OSPA}, the first term, known as the location error, measures track accuracy while the second term, the so called cardinality error, represents the error for missed or false tracks. In this metric, the parameter $p$ controls the sensitivity to
outlier estimates that are distant from the true targets. Moreover,
the metric penalizes the cardinality error by the cut-off parameter
$c$ as a higher value for this parameter penalizes the false and
missing targets more. At a first glance, the selection of the
parameters $c$ and $p$ seems to be critical for the performance of
different methods. However, it was proved that the relation between
the performance of different methods is independent of the
parameters $c$ and $p$ and different values for these parameters only change the scale of
the error, and do not affect the method ranking~\cite{schuhmacher2008consistent}.

\subsubsection{OSPA-T}

Although the OSPA metric measures the multi-target tracking errors such as track accuracy and missed or false tracks, it does not evaluate some other errors such as inconsistent labeling (label switching between the targets in crossing cases) and incorrect label initiation in the case of track truncation. An extension of the OSPA metric, known as OSPA-T\footnote{Contrary to the OSPA, the OSPA-T is not a metric in the formal mathematical sense. However, it can be used as a performance measure.}, considers the aforementioned
errors by measuring the distance between two sets of labeled points. 

Let us assume that we have two labeled sets such that $X=\left\{(l_{1},x_{1}),...,(l_{m},x_{m})\right\}$ and
$Y=\left\{(\breve{l}_{1},y_{1}),...,(\breve{l}_{n},y_{n})\right\}$. In a multi-target tracking problem, these would be the ground truth and estimated state of the targets with their labels in each time frame. In this performance measure, the label correspondence between the estimated and ground truth sets is first performed based on an optimal global assignment using the trajectories' temporal information~\citep{ristic2011metric}. Then, $d^c_p(x,y)$ in Eq.~\ref{eq:OSPA} is substituted by the following distance. 
\begin{align}\label{eq:L-OSPA}
d^{(c,\ell)}_p&\left((l,x),(\breve{l},y)\right)= \\&\min\left(c,\left(\|x-y\|^{p}_p+\left(\ell F_{\delta}(l,\breve{l})\right)^p\right)^{1/p}\right)\nonumber,
\end{align}
where $F_{\delta}(l,\breve{l}) = 0$ if $l$ and $\breve{l}$ are the corresponding labels, and $F_{\delta}(i,j) = 1$ otherwise.
The parameter $\ell$ should be between $0$ and $c$ and controls the penalty assigned to the labeling error. The case $\ell = 0$ assigns no penalty, and $\ell = c$ assigns the maximum penalty. In this paper, we reported the results with $\ell = c$.

\subsection{Evaluation on synthetic data}

\begin{figure}[t]
\begin{minipage}[b]{.49\linewidth}
  \centering
  \centerline{\includegraphics[height=4cm]{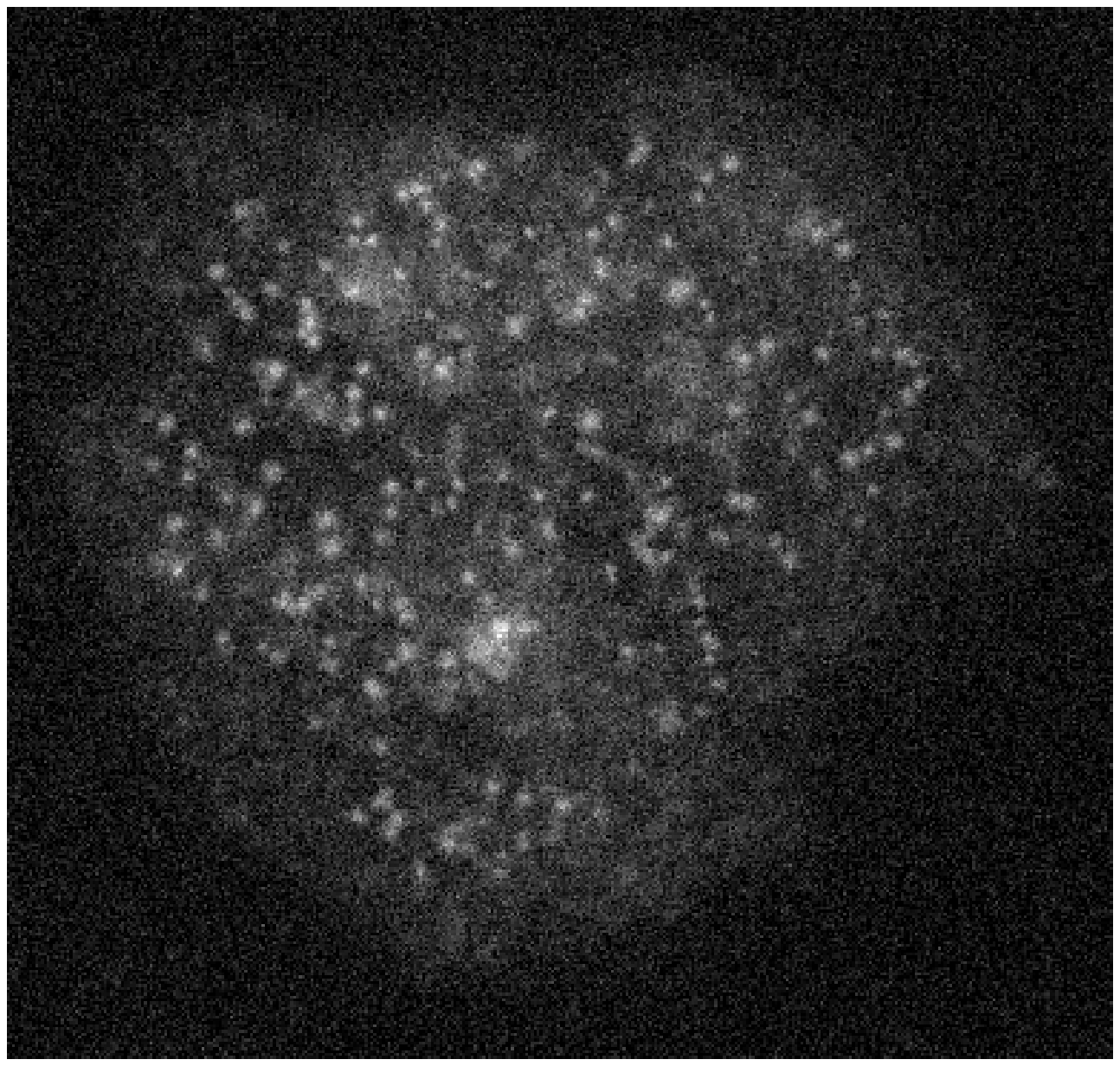}}
  \centerline{(a)}\medskip
\end{minipage}
\hfill
\begin{minipage}[b]{0.49\linewidth}
  \centering
  \centerline{\includegraphics[height=4cm]{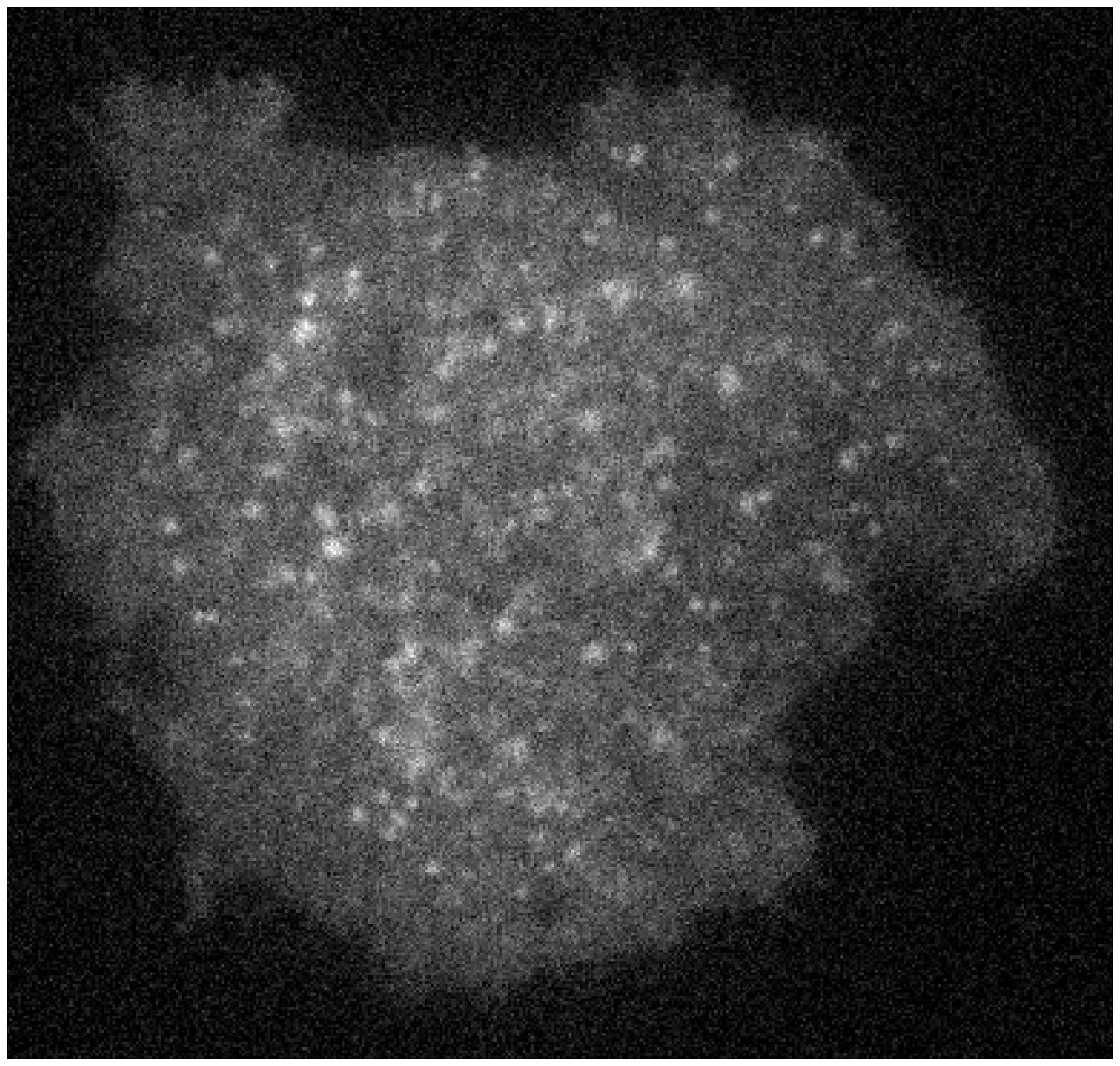}}
  \centerline{(b)}\medskip
\end{minipage}
 \caption{Two frames of the synthetic TIRFM sequences generated by the framework proposed in~\cite{rezatofighi2013framework}.} \label{fig:Synthetic TIRFM}

\end{figure}

To quantitatively evaluate the tracking algorithms, they were first evaluated using $10$ realistic
synthetic movies generated by the framework proposed in~\cite{rezatofighi2013framework}. The sequences simulated using
this framework reflect the difficulties existing in
real TIRFM sequences while providing accurate ground truth. In these sequences, the aim is to mimic the effect of the injection of a stimulus such as insulin into a pancreatic cell. Fig.~\ref{fig:Synthetic TIRFM} shows two frames of the synthetic sequences using this framework which are comparable with the real TIRFM sequences shown in Fig.~\ref{fig:TIRFM}.

Each synthetic sequence was simulated with spatial resolution of $158$nm/pixel and temporal
resolution $10$ fps and consists of time-varying number of targets (on average $\approx190$ particles per frame) moving through $60$ time frames inside a cell
membrane (an estimated background) that is extracted from real TIRFM
sequences with effective region $\approx230\times230$ pixels. The spots were generated in different sizes, $\approx1.2-4.5$ pixels ($200-700$ nm). The dynamics of the targets were
modeled using random walk and linear movement. Furthermore, targets can switch between these two dynamics. The number of intersecting and touching spots in each frame was counted according to the Rayleigh resolution~\cite{rayleigh1879}. The averaged percentage of intersecting and touching spots per frame in these sequences is equal to $1.2\%$.

Due to the 3-D motion of the structures, their intensity
changes according to the TIRFM exponential equation~\cite{rezatofighi2013framework}. Therefore, they may either temporarily or
permanently disappear from the frames. New born targets may also
gradually appear from the background. The sequences are contaminated with Poisson noise and the main characteristics of the sequences such as background intensity and noise level gradually change based on a model extracted from real sequences. Due to spatio-temporally varying noise level and backgrounds and dynamic
intensity of spots, the signal-to-noise ratio (SNR)\footnote{SNR value is used for representing the level of noise and is calculated using definition of Smal \emph{et al.}~\cite{smal2008particle} which is the difference
in intensity between the object and the background, divided by
the standard deviation of the object noise.} of an object cannot be constant. Instead, it varies between $1$ and $9$.

\begin{figure}[t]
   \centerline{\includegraphics[height=6cm,width=8cm]{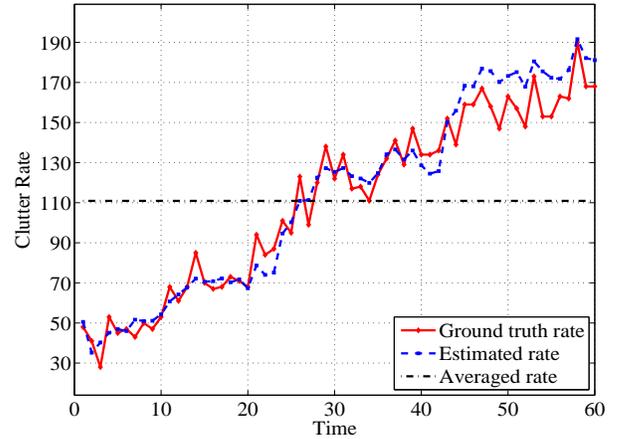}}
 \caption{The ground truth, averaged and estimated (using the proposed
 MM-$\lambda$-$p_D$-CPHD filter) clutter rates for a synthetic image sequence in the high clutter scenario ($\bar{p}_D=0.88$ and $\bar{\lambda}=112$).}
 \label{fig:Estimated_clutter}
 \end{figure}

Similar to the real TIRFM data, the spots in the synthetic sequences fade over time as noise level and background intensity gradually increases. From a biological perspective, it is important to assess the dynamics of vesicles after injection of the stimulus which leads the escalation of noise level and background intensity. Therefore, the threshold in the detection method needs to be set low to ensure consistent detections of the objects in order to avoid early track termination. To this end, we chose a value for the threshold such that the averaged detection probability for all sequences $\bar{p}_D$ is about $0.9$. However, this scenario noticeably increases the clutter rate and its variation over time. Fig.~\ref{fig:Estimated_clutter} shows that the mean clutter rate estimated using the proposed MM-$\lambda$-$p_D$-CPHD filter can appropriately follow the
 quick changes in the ground truth clutter rate.
 
In this experiment, the performance of the trackers was evaluated using these time-varying and highly cluttered detections and their results are reported in Table~\ref{table:OSPA}.  In this Table, the errors are averaged over the number of frames in all $10$ synthetic sequences. The results show that our bootstrap filter benefits from a reliable estimator, the MM-$\lambda$-$p_D$-CPHD filter, which accurately estimates the detection probability and clutter rate. Consequently compared to the other tracker, its tracking results contain fewer false and missing tracks, which decreases the cardinality error significantly. Similarly, this error is also low in the MM-$\lambda$-$p_D$-CPHD filter. The other Bayesian trackers cannot benefit from these estimations as their parameters are fixed. Therefore, they have higher cardinality error. In comparison with traditional Bayesian trackers such as the IMM-JPDA and MHT trackers, the RFS filters have relatively better cardinality error as they properly incorporate birth,
 death and clutter models in their formulations. However, the higher difference between the OSPA and OSPA-T errors of the all RFS filters represent that their results include more identity switch errors compared to the other trackers due to the simple tag propagation scheme used. 
 
 The performance of the P-tracker as one of the deterministic trackers seems to be sensitive to cluttered measurements. Therefore, it has the highest cardinality error of the trackers. In contrast, the results show that another deterministic tracking scheme, U-tracker, can robustly track the particles while dealing with highly cluttered detections. 
 
The P-Tracker has the lowest location error of all trackers.  Although this error reflects the accuracy of the trackers in tracking particles, its lower value can be also an artifact of the tracker's poor
performance. The trackers with higher cardinality error have relatively lower location error and vice versa.
 
 {\renewcommand{\arraystretch}{1.5}%
 \begin{table}[b]
 \caption{The averaged location, cardinality, OSPA and OSPA-T errors of the trackers in pixel with the metric parameters $p=1$ and $c=\ell=10$ (pixel) for the $10$ synthetic sequences in the high clutter scenario ($\bar{p}_D=0.88$ and $\bar{\lambda}=112$).}
 \centering 
 \begin{tabular}{|c|| cc c| c|}
 \hline Method &Location&Cardinality&OSPA&OSPA-T \\
 \hline\hline 
 B-MM-CPHD&$2.52$&$\mathbf{0.56}$&$\mathbf{3.08}$&$5.41$\\
 \hline
 MM-$\lambda$-$p_{D}$-CPHD &$2.64$&$0.63$&$3.27$&$5.48$\\
 \hline
 MM-CPHD&$2.39$&$1.12$&$3.51$&$5.50$\\
 \hline
 MM-PHD~\cite{rezatofighi2013multiple}&$2.71$&$1.30$&$4.01$&$6.55$\\
 \hline
 IMM-JPDA~\cite{rezatofighi2012application}&$1.54$&$2.68$&$4.22$&$5.72$\\
 \hline
MHT~\cite{chenouard2013multiple}&$1.71$&$2.23$&$3.94$&$5.67$\\
 \hline
P-Tracker~\cite{sbalzarini2005feature}&$\mathbf{1.33}$&$2.70$&$4.03$&$6.50$\\
 \hline
U-Tracker~\cite{jaqaman2008robust}&$1.85$ &$1.43$ & $3.28$&$\mathbf{5.04}$\\
 \hline
 \end{tabular}
 \label{table:OSPA}
 \end{table}
 
The method rankings may change in other scenarios, e.g. where the clutter rate is low and its variation over time can be ignored. In our application, the clutter rate and its variation over time can be decreased by increasing the detection threshold value. For example, we chose a value such that the averaged clutter rate $\bar{\lambda}$ and detection probability $\bar{p}_D$ are respectively about $11$ and $0.7$. In this case, the detected points from a target are inconsistent and many faint particles are not detected after initial time frames. This case is undesirable for our application due to many truncated tracks with short life time. Nevertheless, we report the tracker's OSPA and OSPA-T errors in this case in Table~\ref{table:OSPA_low clutter} in order to compare their different performance in the low and high clutter rate scenarios. Note that both false tracks in high clutter scenario and missed or truncated tracks in low clutter scenario increase the OSPA and OSPA-T errors.} 
  
    \begin{figure}[t]
      \centerline{\includegraphics[height=6cm,width=8cm]{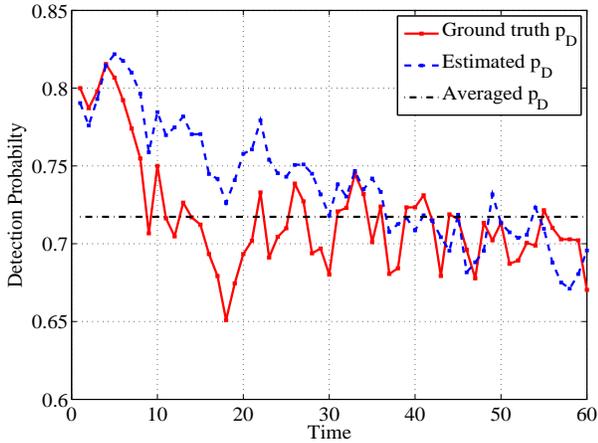}}
    \caption{The ground truth, averaged and estimated (using the proposed
    MM-$\lambda$-$p_D$-CPHD filter) detection probability for a synthetic image sequence in the low clutter scenario ($\bar{p}_D=0.7$ and $\bar{\lambda}=11$).}
    \label{fig:Estimated_PD}
    \end{figure}
    
     {\renewcommand{\arraystretch}{1.5}%
     	\begin{table}[b]
     		\caption{The averaged location, cardinality, OSPA and OSPA-T errors of the trackers in pixel with the metric parameters $p=1$ and $c=\ell=10$ (pixel) for the $10$ synthetic sequences in the low clutter scenario ($\bar{p}_D=0.7$ and $\bar{\lambda}=11$).}
     		\centering 
     		\begin{tabular}{|c|| cc c| c|}
     			\hline Method &Location&Cardinality&OSPA&OSPA-T \\
     			\hline\hline 
     			B-MM-CPHD&$2.71$&$\mathbf{0.68}$&$3.39$&$5.39$\\
     			\hline
     			MM-$\lambda$-$p_{D}$-CPHD &$2.78$&$0.82$&$3.60$&$5.97$\\
     			\hline
     			MM-CPHD&$2.84$&$0.79$&$3.63$&$5.61$\\
     			\hline
     			MM-PHD~\cite{rezatofighi2013multiple}&$2.90$&$1.08$&$3.98$&$6.01$\\
     			\hline
     			IMM-JPDA~\cite{rezatofighi2012application}&$1.37$&$1.67$&$\mathbf{3.04}$&$\mathbf{4.21}$\\
     			\hline
     			MHT~\cite{chenouard2013multiple}&$0.81$&$3.07$&$3.88$&$5.48$\\
     			\hline
     			P-Tracker~\cite{sbalzarini2005feature}&$0.85$&$2.55$&$3.40$&$5.20$\\
     			\hline
     			U-Tracker~\cite{jaqaman2008robust}&$\mathbf{0.64}$ &$3.36$ & $4.00$&$4.93$\\
     			\hline
     		\end{tabular}
     		\label{table:OSPA_low clutter}
     	\end{table}

    The results show that the trackers such as the IMM-JPDA filter and P-Tracker which have the worst performance in highly cluttered measurements perform better for a low clutter rate with long missed detections compared to other trackers. In contrast, the U-Tracker cannot perform reliably in this case as it has the highest cardinality error. Estimating declining trend in the detection probability (Fig.~\ref{fig:Estimated_PD}) helps our bootstrap tracker to still have lower OSPA error compared to the other trackers, but relatively high OSPA-T error due to the simple tag propagation scheme. 
    
    In a nutshell, the performance of the trackers varies according to the detector reliability. Our bootstrap approach has superior performance for highly cluttered detections with the time-varying rates.

\subsection{Evaluation on real data}
The tracking methods were also tested on a real TIRFM
sequence with spatial and temporal resolution of $\approx150$nm/pixel and $10$ fps respectively. The image sequences were acquired from a pancreatic beta cell injected by insulin during the acquisition.  

In order to prepare a ground truth, an independent expert manually tracked all visible structures ($332$ trajectories from $21752$ spots) in a part of the cell within $500$ time frames using the freely available software MTrackJ~\cite{meijering2006mtrackj}. The annotated trajectories were double-checked by another biologist to ensure reliability of the ground truth.
   
In order to detect and track all structures, especially after injection of insulin, the threshold in the detection method was set low, which
noticeably increases clutter rate and its variation over time (Fig.~\ref{fig:Estimated_clutter_detection_real}). The results of the trackers for the real data are reported in Table ~\ref{table:OSPA real}. In this Table, the errors are averaged over the number of time steps in the real image sequence. 

 \begin{figure*}[t]
 	\begin{minipage}[b]{.49\linewidth}
 		\centering
 		\centerline{\includegraphics[height=6cm,width=9cm]{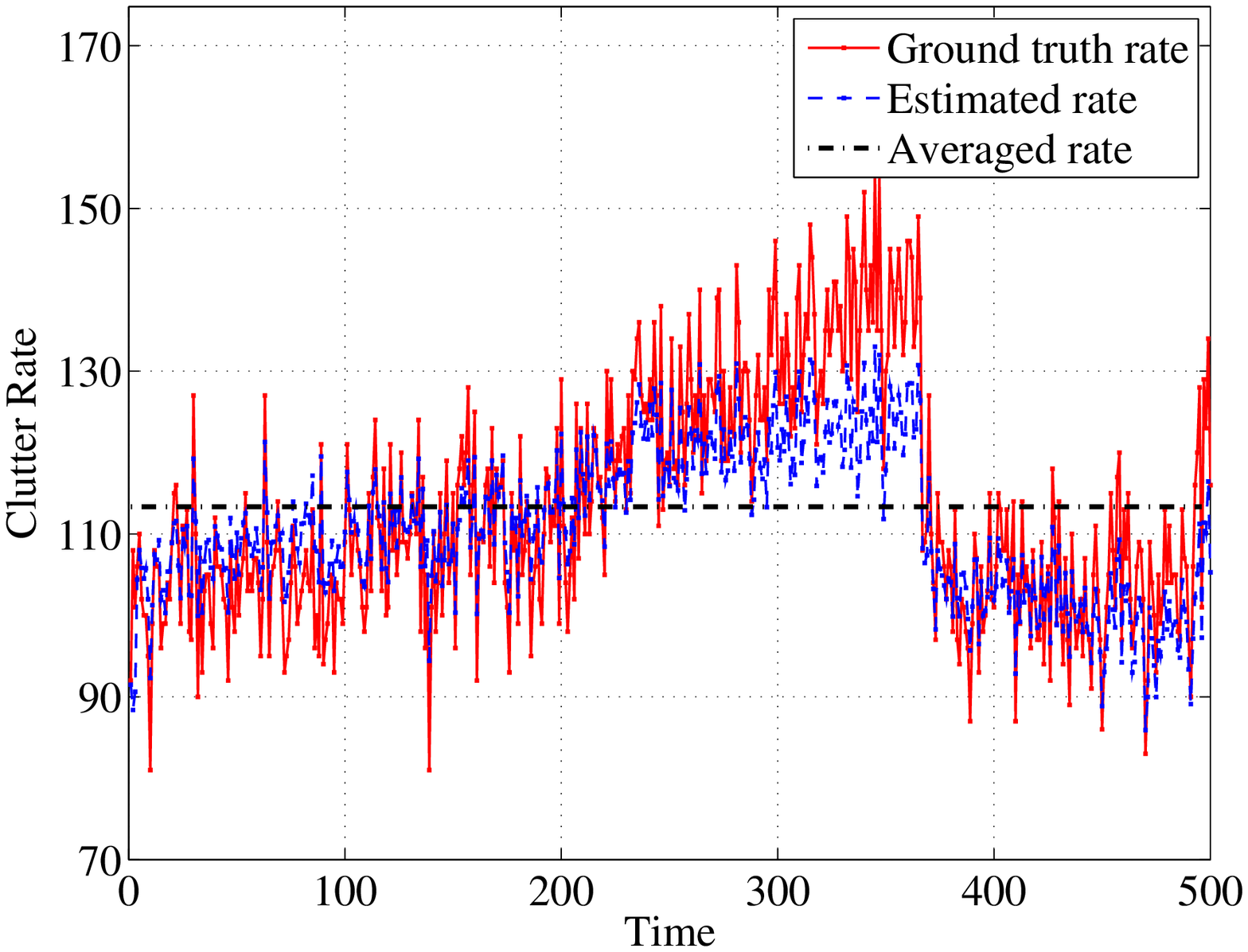}}
 		\centerline{ (a)}\medskip
 	\end{minipage}
 	\begin{minipage}[b]{.49\linewidth}
 			\centering
 			\centerline{\includegraphics[height=6cm,width=9cm]{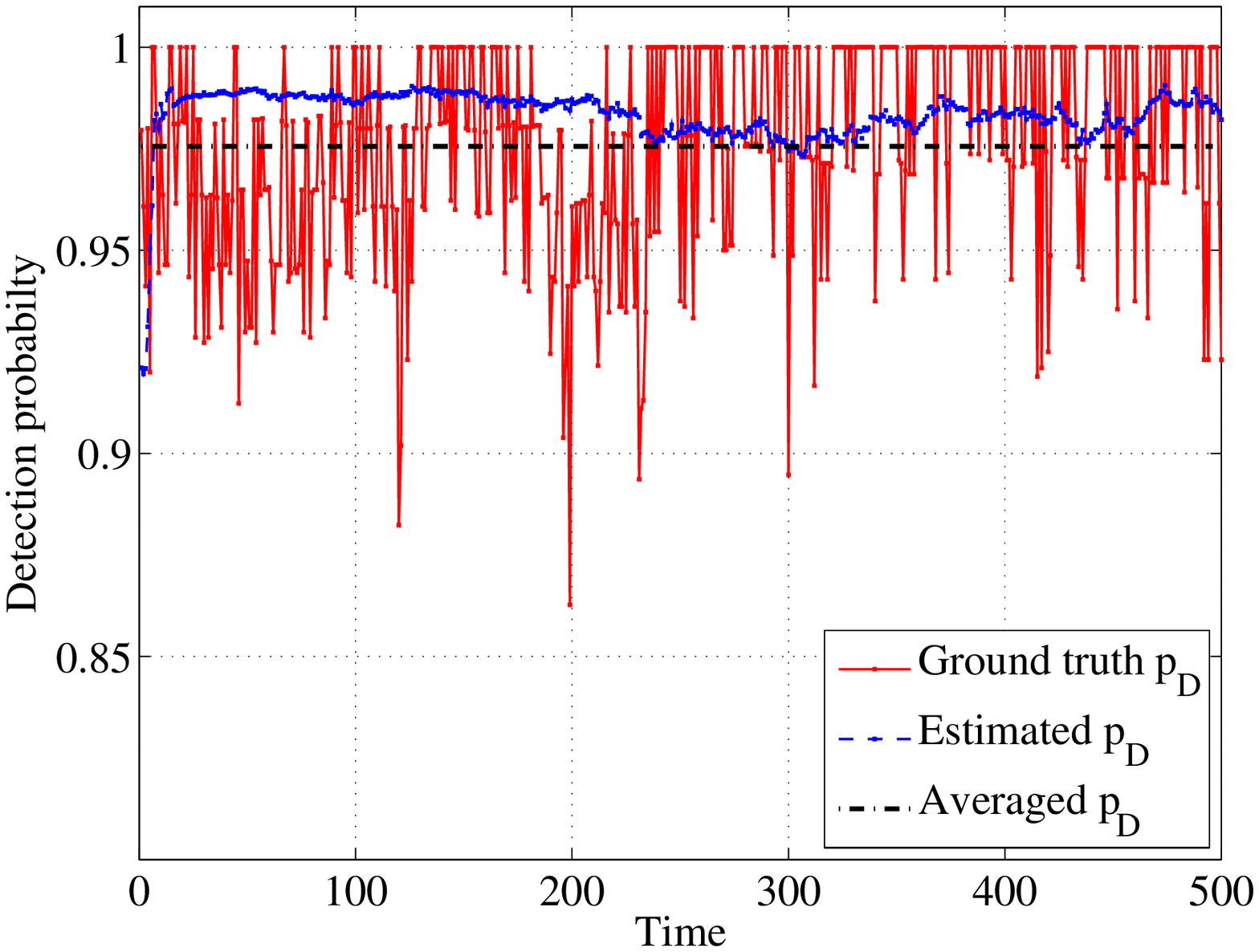}}
 			\centerline{ (b)}\medskip
 	\end{minipage}

 	\caption{The ground truth, averaged and estimated (a) clutter rate and (b) detection probability using the proposed
 			MM-$\lambda$-$p_D$-CPHD filter for the real image sequence.}
 	\label{fig:Estimated_clutter_detection_real}
 \end{figure*}

The higher clutter rate along with significantly higher number of time frames and numerous faint structures existing in real sequences cause higher values of the OSPA and OSPA-T errors of the all trackers for real sequences compared to the synthetic data. In addition, the results confirm our arguments about the performance of the trackers in the synthetic sequences for the same high clutter scenario. The ability of our B-MM-CPHD to properly track true targets while accurately estimating the clutter rate and detection probability (Fig.~\ref{fig:Estimated_clutter_detection_real}) leads to the lowest cardinality and also OSPA errors.

Since there were always faint structures in real sequences that even experienced annotators were unable to detect or determine whether they are real or false targets, the reliability of the manual ground truth cannot be completely guaranteed. In this case, the quantitative results may be biased to a specific method. To maximize the validity of our comparison, the results of the tracking were also visually assessed by the
experts. This assessment also showed that our bootstrap filter can
better detect and track the real vesicles, especially faint ones, while avoiding tracking
false targets. 

In Fig.~\ref{fig:Tracking results} (a), the tracking results of
several crossing particles with maneuvering motions using the proposed
bootstrap filter are shown. The results show that it can
properly deal with the maneuvering motion of the targets. However,
the results are not error free and include some missing and false tracks and switching labeling errors. Fig.~\ref{fig:Tracking results} (b) also demonstrates the performance of our proposed bootstrap filter in tracking two faint particles moving through the synthetic and real sequences with time-varying background and noise
level.

{\renewcommand{\arraystretch}{1.5}%
\begin{table}[b]
\caption{The averaged location, cardinality, OSPA and OSPA-T errors of the trackers} in pixel with the metric parameters $p=1$ and $c=\ell=10$ (pixel) for a real image sequence.
\centering 
\begin{tabular}{|c|| cc c| c|}
\hline Method &Location&Cardinality& OSPA& OSPA-T \\
\hline\hline 
B-MM-CPHD&$3.91$&$\mathbf{1.04}$&$\mathbf{4.95}$&$\mathbf{6.30}$\\
\hline
MM-$\lambda$-$p_D$-CPHD &$4.01$&$1.23$&$5.24$&$6.56$\\
\hline
MM-CPHD&$3.92$&$2.07$&$5.99$&$7.23$\\
\hline
MM-PHD~\cite{rezatofighi2013multiple}&$4.05$&$2.23$&$6.28$&$7.64$\\
\hline
IMM-JPDA~\cite{rezatofighi2012application}&$0.93$&$5.75$&$6.68$&$7.25$\\
\hline
MHT~\cite{chenouard2013multiple}&$1.21$ & $4.28$& $5.49$&$6.33$\\
\hline
P-Tracker~\cite{sbalzarini2005feature}&$\mathbf{0.55}$ & $5.45$ & $6.00$&$7.32$\\
\hline
U-Tracker~\cite{jaqaman2008robust}& $1.96$ & $3.63$ & $5.59$&$6.44$\\
\hline
\end{tabular}
\label{table:OSPA real}
\end{table}

\begin{figure*}[thb]
\begin{minipage}[b]{.29\linewidth}
  \centering
  \centerline{\includegraphics[height=4.8cm]{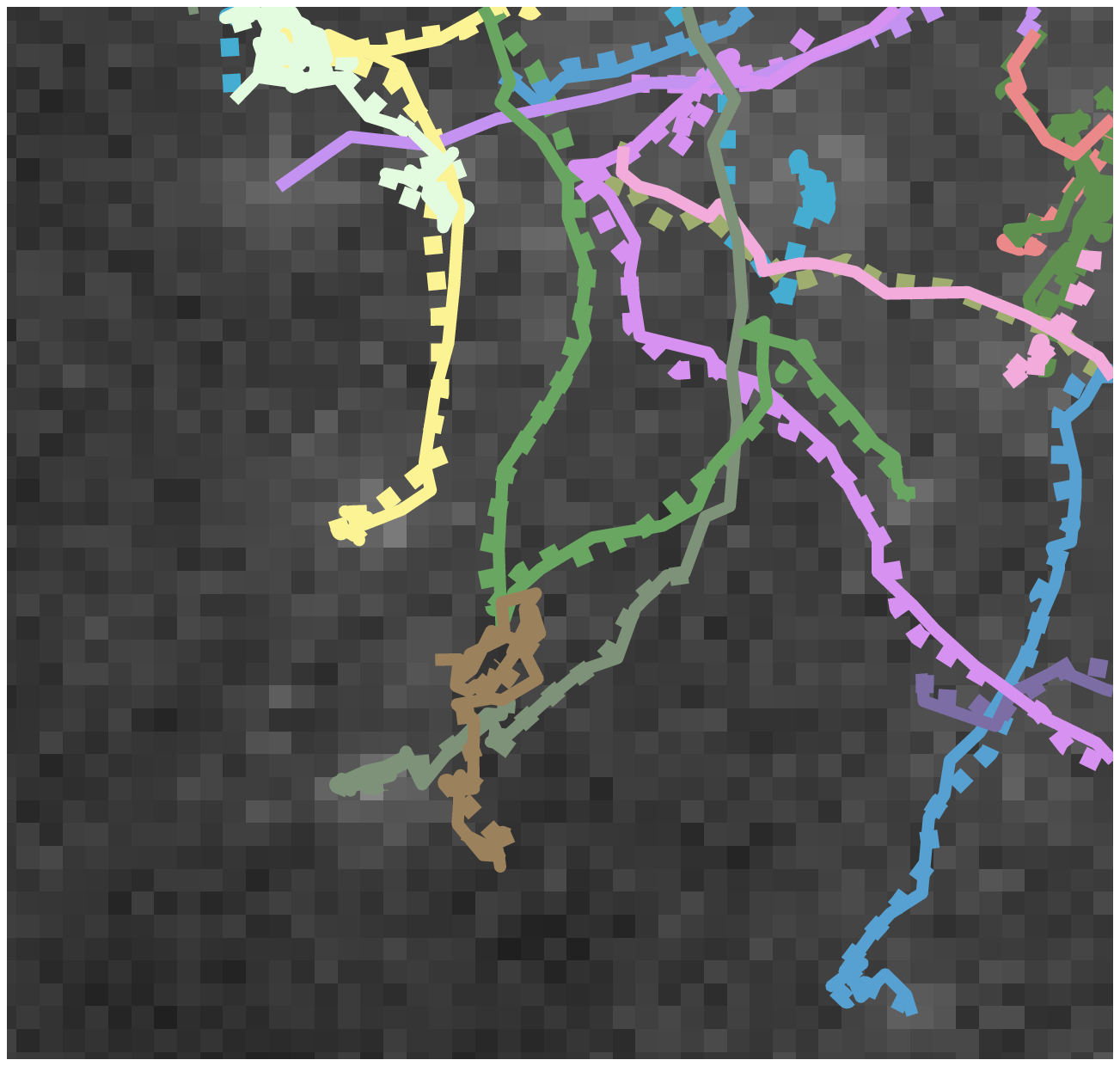}}
    \centerline{(a)}\medskip
\end{minipage}
\begin{minipage}[b]{.71\linewidth}
\begin{minipage}[b]{.19\linewidth}
  \centering
  \centerline{\includegraphics[height=2.4cm]{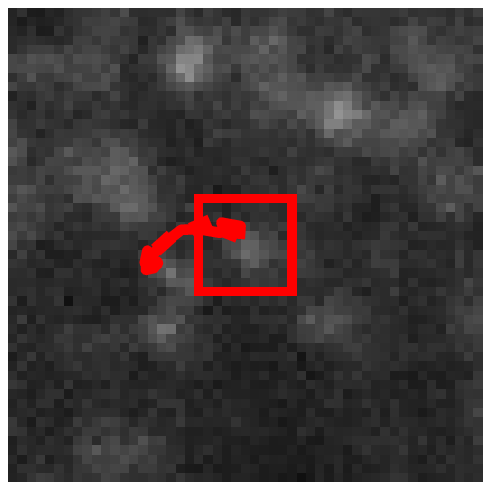}}
\end{minipage}
\hfill
\begin{minipage}[b]{0.19\linewidth}
  \centering
  \centerline{\includegraphics[height=2.4cm]{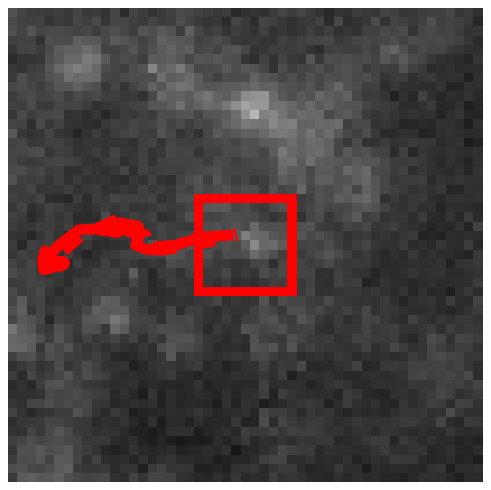}}
\end{minipage}
\hfill
\begin{minipage}[b]{0.19\linewidth}
  \centering
  \centerline{\includegraphics[height=2.4cm]{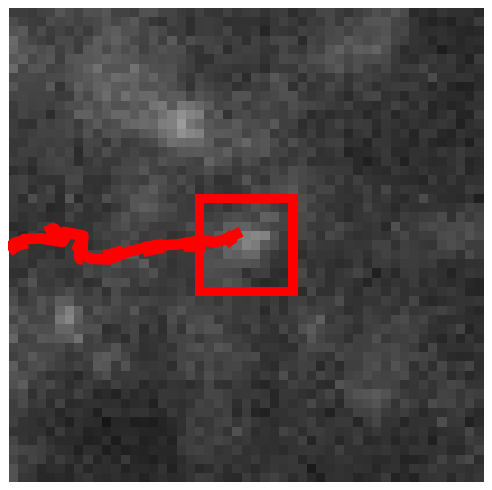}}
\end{minipage}
\hfill
\begin{minipage}[b]{0.19\linewidth}
  \centering
  \centerline{\includegraphics[height=2.4cm]{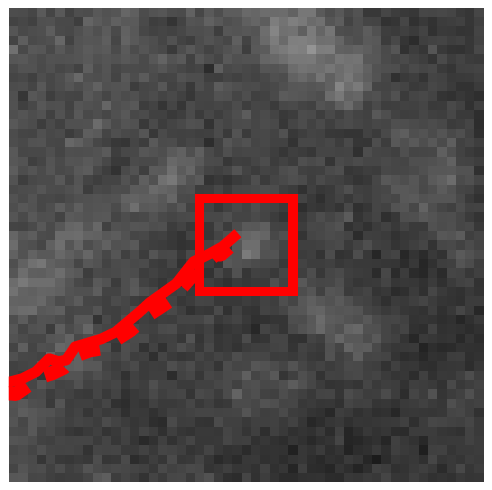}}
\end{minipage}
\hfill
\begin{minipage}[b]{0.19\linewidth}
  \centering
  \centerline{\includegraphics[height=2.4cm]{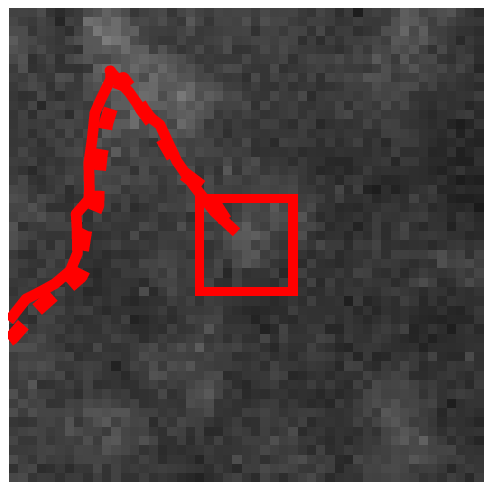}}
\end{minipage}

\begin{minipage}[b]{0.19\linewidth}
  \centering
  \centerline{\includegraphics[height=2.4cm]{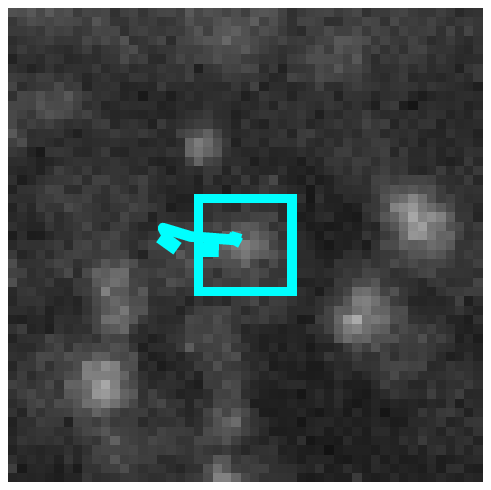}}
\end{minipage}
\hfill
\begin{minipage}[b]{0.19\linewidth}
  \centering
  \centerline{\includegraphics[height=2.4cm]{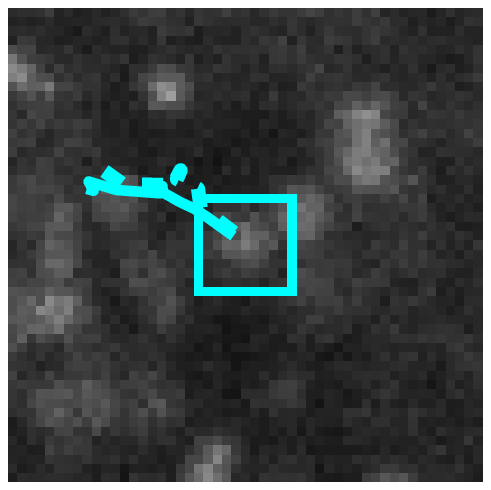}}
\end{minipage}
\hfill
\begin{minipage}[b]{0.19\linewidth}
  \centering
  \centerline{\includegraphics[height=2.4cm]{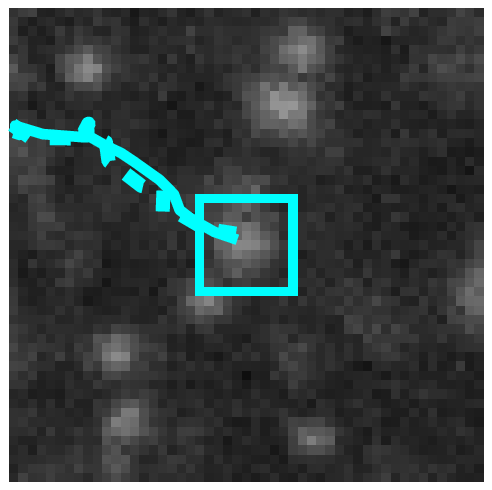}}
\end{minipage}
\hfill
\begin{minipage}[b]{0.19\linewidth}
  \centering
  \centerline{\includegraphics[height=2.4cm]{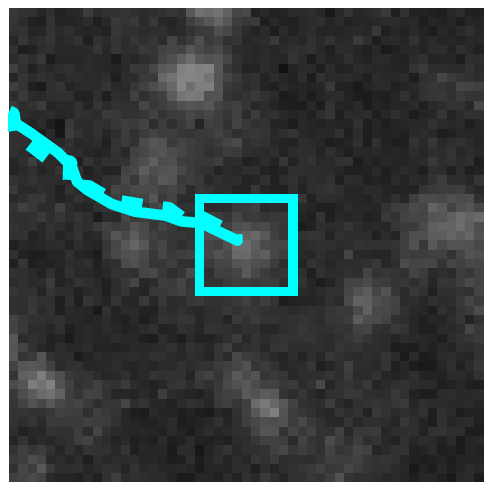}}
\end{minipage}
\hfill
\begin{minipage}[b]{0.19\linewidth}
  \centering
  \centerline{\includegraphics[height=2.4cm]{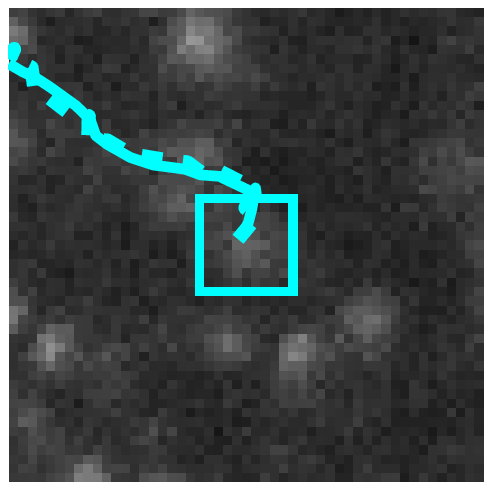}}
\end{minipage}
  \centerline{(b)}\medskip
\end{minipage}
 \caption{Some examples of tracked vesicles using the
proposed bootstrap filter (dashed line) against the ground truth
(solid line) in the synthetic and real TIRFM sequences. (a) The
tracking results of multiple crossing particles with maneuvering
motions in the synthetic TIRFM sequences. (b) The resulting
trajectory of two faint vesicles from the synthetic (Top row) and real
(Bottom row) TIRFM sequences in different time frames. For an
improved visualization, the other tracks are eliminated. These
particles move in different background intensity and noise level.} \label{fig:Tracking results}
\end{figure*}
\section{Conclusion}

The estimation of clutter rate and detection probability helps improve tracking of the targets in particle tracking applications where measurements include many spurious and missed detections with unknown and time-varying parameters.  The RFS framework provides a principled solution to deal with the estimation of these parameters which was not possible in previous approaches. The $\lambda $-$p_D$-CPHD filter is one of the filters derived based on RFS theory which is able to estimate these parameters. However, the filter cannot naturally perform as well as the  CPHD filter
with known parameters. To this end, in this paper we proposed a bootstrap filter by combination of the CPHD and $\lambda $-$p_D$-CPHD filters. To accommodate the maneuvering motion, we also proposed a multiple model implementation of the filters. Therefore, the clutter rate and detection probability are estimated by the multiple model $\lambda $-$p_D$-CPHD filter bootstrapped
onto a multiple model CPHD filter that outputs target estimates.

The proposed approach was evaluated on a challenging particle tracking application where vesicles move in noisy sequences of total internal reflection fluorescence microscopy while the noise
characteristic and background intensity of the sequences
change during the acquisition process. In this application which the clutter rate and detection probability are time-varying, we demonstrated that our bootstrap filter is able to better track the real targets and more reliably avoid false tracks compared to the other RFS filters
such as the MM-PHD, MM-CPHD and MM-$\lambda$-$p_D$-CPHD as well
as other state-of-the-art particle tracking methods such as the IMM-JPDA, MHT, P-Tracker and U-Tracker in both synthetic and real sequences. However, in the applications where the motion of each individual targets is required, the tag propagation scheme may need to be changed to a track management algorithm, e.g., that proposed in~\cite{panta2007novel,lin2006track} or more generally, the tracker in the bootstrap filter can be replaced by any multi-target tracker that requires knowledge of false alarm and detection rate and performs reliably in the label assignments. In the RFS concept, a principled solution to the track labeling problem using labeled RFS~\cite{vo2013labeled} has been recently proposed which can be also used for this purpose.    

Obviously, our approach may not be a good choice for some applications. For example, in the cases where the particles can be easily detected or the rates for false alarms and missed detections are negligible, known or time-invariant, there is no point to adaptively estimate the clutter rate and the detection profile. In addition, to estimate these time-varying rates, the MM-$\lambda$-$p_D$-CPHD requires some probabilistic priors describing how the clutter rate and the detection probability change over time~\cite{mahler2011cphd}. Therefore, this approach requires more parameters compared to other trackers. Moreover, the bootstrap method filters twice in each frame in order to estimate the state of the targets. Consequently, its processing time is more than each of its component filters, MM-$\lambda$-$p_D$-CPHD and MM-CPHD. Our non-optimized MATLAB code was run on an ordinary PC (Intel Core $2$ Quad, $2.66$ GHz CPU, $8$ GB RAM). The average CPU processing time per frame per target in the highest considered clutter rate (around $280$ detected positions per frame) for the multiple model CPHD, $\lambda$-$p_D$-CPHD and our bootstrap filter are about $17.1$ms, $8.0$ms and $25.1$ms respectively.

Due to linearity of the dynamic and measurement models in our application, we only used the multiple model Gaussian linear implementation of all filters. The recursive equations derived in the appendices can be used for both Gaussian linear and non-Gaussian non-linear system models. In order to apply the proposed framework to applications with non-linear non-Gaussian system models, the sequential Monte Carlo (SMC) implementation of the filters is required. This implementation will also allows us to further evaluate the performance of our bootstrap filter and compare it against the SMC-PHD~\cite{vo2005sequential},  SMC-CPHD~\cite{mahler2007statistical} and the traditional particle~\cite{hue2002tracking} filters. 

\appendices
\section*{Appendix}
For completeness, we now derive the recursive equations for both the
multiple model CPHD and $\lambda$-$p_D$-CPHD filters in this section.
The following notations are used in throughout this Appendix. The
binomial and permutation coefficients are denoted by $C^l_j$ and
$P^n_j$, respectively. $\langle \cdot,\cdot\rangle$ is the inner
product operation between two continuous or two discrete functions.
The elementary symmetric function of order $j$ defined for a finite
set $Z$ of real numbers is denoted by 
\begin{equation}
e_j(Z)=\sum_{S\subseteq
Z,|S|=j}\left(\prod_{\theta\in S}\theta\right),
\end{equation}
where $|S|$ is the
cardinality of a set $S$ and $e_0(Z)=1$~\cite{vo2007analytic}.
\renewcommand\thesubsection{\thesection\Alph{subsection}}
\subsection{Multiple Model CPHD Recursions} \label{appendix:CPHD}
 \emph{Prediction step:} Suppose at time $k-1$, the posterior
cardinality distribution $\rho_{k-1}$ and posterior intensity
$\underline{v}_{k-1}$ are known. The predicted cardinality
distribution $\rho_{k|k-1}$ and predicted intensity
$\underline{v}_{k|k-1}$ are calculated by 
\begin{align}\label{eq:predicted cardinality}
 \rho_{k|k-1}(n)= \sum_{j=0}^n
 \rho_{\Gamma}(n-j)\Pi[\underline{v}_{k-1},\rho_{k-1}](j),
\end{align}

\begin{equation}\label{eq:predicted intensity}
\begin{aligned}
\underline{v}_{k|k-1}&(x,r)=\gamma(x)\pi(r)+\\
&\sum_{\acute{r}}\!\!\int\!\!
\underline{p}_{S}(\acute{x},\acute{r})f(x|\acute{x},r)\tau(r|\acute{r})
\underline{v}_{k-1}(\acute{x},\acute{r})d\acute{x},
\end{aligned}
\end{equation}
where $\Pi[\underline{v}_{k-1},\rho_{k-1}](j)=$
\begin{equation}
\sum_{l=j}^{\infty}C^l_j\rho_{k-1}(l)\frac{\langle
\underline{p}_{S},\underline{v}_{k-1}\rangle^j\langle1-\underline{p}_{S},\underline{v}_{k-1}\rangle^{l-j}}{\langle1,\underline{v}_{k-1}\rangle^{l-j}}.
\end{equation}
Note that in the multiple model approach, the inner product function
operates on both the kinematic state and the model, i.e. $\langle
\underline{p}_{S},\underline{v}_{k-1}\rangle=\sum_{\acute{r}}\int
p_{S}(\acute{x},\acute{r})\underline{v}_{k-1}(\acute{x},\acute{r})d\acute{x}$.

 \emph{Update step:} If at time $k$, the predicted cardinality
distribution $\rho_{k|k-1}$, predicted intensity
$\underline{v}_{k|k-1}$ and set of measurement $Z_k$ are given, the
updated cardinality distribution $\rho_{k}$ and updated intensity
$\underline{v}_{k}$ are calculated by 
\begin{align} \label{eq:updated cardinality}
\rho_k(n)=\frac{\underline{\Upsilon}^0
\left[\underline{v}_{k|k-1},Z_k\right](n)\rho_{k|k-1}(n)}{\left\langle\underline{\Upsilon}^0
\left[\underline{v}_{k|k-1},Z_k\right],\rho_{k|k-1}\right\rangle},
\end{align}
\begin{multline}
\underline{v}_{k}(x,r)=\underline{v}_{k|k-1}(x,r)\times\\
\left[(1-\underline{p}_{D}(x,r))\frac{\left\langle\underline{\Upsilon}^1
\left[\underline{v}_{k|k-1},Z_k\right],\rho_{k|k-1}\right\rangle}{\left\langle\underline{\Upsilon}^0
\left[\underline{v}_{k|k-1},Z_k\right],\rho_{k|k-1}\right\rangle}\right.\\+
\left.\sum_{z\in
Z_k}\underline{\psi}_{z}(x,r)\frac{\left\langle\underline{\Upsilon}^1
\left[\underline{v}_{k|k-1},Z_k\backslash{z}\right],\rho_{k|k-1}\right\rangle}{\left\langle\underline{\Upsilon}^0
\left[\underline{v}_{k|k-1},Z_k\right],\rho_{k|k-1}\right\rangle}\right],
\end{multline}
where $\underline{\Upsilon}^u
\left[\underline{v}_{k|k-1},Z_k\right](n)=$
\begin{equation}
\begin{aligned}
&\sum_{j=0}^{\min(|Z_k|,n)}(|Z_k|-j)!\rho_{K}(|Z_k|-j)P^n_{j+u}
\\
&\times\frac{\langle
1-\underline{p}_{D},\underline{v}_{k|k-1}\rangle^{n-(j+u)}}{\langle
1,\underline{v}_{k|k-1}
\rangle^n}e_j\left(\Xi(\underline{v}_{k|k-1},Z_k)\right)
\end{aligned}
\end{equation}
where
\begin{equation}
\Xi(\underline{v}_{k|k-1},Z_k)=\left\{\left\langle\underline{v}_{k|k-1},\underline{\psi}_{z}\right\rangle:
z\in Z_k\right\},
\end{equation}
and 
\begin{equation} 
\displaystyle
\underline{\psi}_{z}(x,r)=\frac{\langle
1,\kappa\rangle}{\kappa(z)}\underline{g}(z|x,r)\underline{p}_{D}(x,r).
\end{equation}

\subsection{Multiple Model $\lambda$-$p_D$-CPHD Recursions}
\label{appendix:lambda-CPHD} \emph{Prediction step:} Suppose at
time $k-1$, the hybrid cardinality distributions $\ddot{\rho}_{k-1}$
and the posterior intensity distribution for actual targets
$\underline{v}^{(1)}_{k-1}$ and clutter generators
$\underline{v}^{(0)}_{k-1}$ are given. The hybrid predicted
cardinality distribution $\ddot{\rho}_{k|k-1}$ and predicted
intensity for actual targets $\underline{v}^{(1)}_{k|k-1}$ and
clutter generators $\underline{v}^{(0)}_{k|k-1}$ are calculated by
\begin{align} \label{eq:lambda-predicted cardinality}
 \ddot{\rho}_{k|k-1}(\ddot{n})= \sum_{j=0}^{\ddot{n}}
 \ddot{\rho}_{\Gamma}(\ddot{n}-j)\sum_{l=j}^{\infty}C^l_j\ddot{\rho}_{k-1}(l)(1-\phi)^{l-j}\phi^j
\end{align}
where 
\begin{equation}
\displaystyle \phi=\left(\frac{\langle
\underline{p}^{(1)}_{S},\underline{v}^{(1)}_{k-1}\rangle+\langle
\underline{p}^{(0)}_{S},\underline{v}^{(0)}_{k-1}\rangle}{\langle1,\underline{v}^{(1)}_{k-1}\rangle+\langle1,\underline{v}^{(0)}_{k-1}\rangle}\right),
\end{equation}

\begin{multline}
\underline{v}^{(1)}_{k|k-1}(x,a,r)=\underline{\gamma}^{(1)}_{k}(x,a)\pi(r)+\\
\sum_{\acute{r}}\int\int_0^1\left[p^{(1)}_{S}(\acute{x},\acute{r})f^{(1)}(x|\acute{x},r)\times\right.\\
\left.f^{(\Delta)}(a|\acute{a},\acute{r})\tau(r|\acute{r})
\underline{v}^{(1)}_{k-1}(\acute{x},\acute{a},\acute{r})d\acute{x}d\acute{a}\right],
\end{multline}
\begin{align}
\underline{v}^{(0)}_{k|k-1}(b)=\underline{\gamma}^{(0)}(b)+p^{(0)}_{S}\underline{v}^{(0)}_{k-1}(b).
\end{align}

\emph{Update step:} If at time $k$, the predicted intensity for
actual targets $\underline{v}^{(1)}_{k|k-1}$, the predicted
intensity for clutter generators $\underline{v}^{(0)}_{k|k-1}$, the
predicted hybrid cardinality distribution $\ddot{\rho}_{k|k-1}$ and
set of measurement $Z_k$ are all given and the function
$\underline{\ddot{\Upsilon}}^u
\left[\underline{\ddot{v}}_{k|k-1},Z_k\right](\ddot{n})$ defined as
follows:
\begin{multline}
\underline{\ddot{\Upsilon}}^u
\left[\underline{\ddot{v}}_{k|k-1},Z_k\right](\ddot{n})=\\
\left\{
\begin{array}{ll}
0&\ddot{n}<|Z_k|+u\\
P^{\ddot{n}}_{|Z_k|+u}\Phi^{\ddot{n}-(|Z_k|+u)}&\ddot{n}\geq|Z_k|+u
\end{array} \right.
\end{multline}
where 
\begin{equation}
\displaystyle
\Phi=1-\frac{\left\langle\underline{v}^{(1)}_{k|k-1},\underline{p}^{(1)}_{D}\right\rangle+\left\langle\underline{v}^{(0)}_{k|k-1},\underline{p}^{(0)}_{D}\right\rangle}{\left\langle
1,\underline{v}^{(1)}_{k|k-1}\right\rangle+\left\langle1,\underline{v}^{(0)}_{k|k-1}\right\rangle},
\end{equation}
where
$\underline{p}^{(1)}_{D}(x,a,r)=a$ and
$\underline{p}^{(0)}_{D}(b)=b$.

The updated cardinality distribution $\ddot{\rho}_k$ and the updated
intensity distribution for actual targets $\underline{v}^{(1)}_k$
and clutter generators $\underline{v}^{(0)}_k$ are given as follows
\begin{align}
\ddot{\rho}_k(\ddot{n})=\left\{
\begin{array}{ll}
0&\ddot{n}<|Z_k|,\\
\frac{ \displaystyle\underline{\ddot{\Upsilon}}^0
\left[\underline{\ddot{v}}_{k|k-1},Z_k\right](\ddot{n})\ddot{\rho}_{k|k-1}(\ddot{n})}{\displaystyle \left\langle\underline{\ddot{\Upsilon}}^0,\ddot{\rho}_{k|k-1}\right\rangle}&
\ddot{n}\geq|Z_k|,
\end{array} \right. 
\end{align}
\begin{multline}
\underline{v}^{(1)}_{k}(x,a,r)=\underline{v}^{(1)}_{k|k-1}(x,a,r)\\
\left[\frac{(1-a)\frac{\left\langle\underline{\ddot{\Upsilon}}^1
\left[\underline{\ddot{v}}_{k|k-1},Z_k\right],\ddot{\rho}_{k|k-1}\right\rangle}{\left\langle\underline{\ddot{\Upsilon}}^0
\left[\underline{\ddot{v}}_{k|k-1},Z_k\right],\ddot{\rho}_{k|k-1}\right\rangle}}{\left\langle
1,\underline{v}^{(1)}_{k|k-1}\right\rangle+\left\langle1,\underline{v}^{(0)}_{k|k-1}\right\rangle}+\right.\\
\left.\sum_{z\in
Z_k}\frac{a.g(z|x,r)}{\left\langle\underline{v}^{(0)}_{k|k-1},p^{(0)}_{D}\mathcal{K}\right\rangle+\left\langle\underline{v}^{(1)}_{k|k-1},p^{(1)}_{D}g(z|\cdot)\right\rangle}\right],
\end{multline}
\begin{multline}\label{eq:clutter-updated intensity}
\underline{v}^{(0)}_{k}(b)=\underline{v}^{(0)}_{k|k-1}(b)\left[\frac{(1-b)\frac{\left\langle\underline{\ddot{\Upsilon}}^1
\left[\underline{\ddot{v}}_{k|k-1},Z_k\right],\ddot{\rho}_{k|k-1}\right\rangle}{\left\langle\underline{\ddot{\Upsilon}}^0
\left[\underline{\ddot{v}}_{k|k-1},Z_k\right],\ddot{\rho}_{k|k-1}\right\rangle}}{\left\langle
1,\underline{v}^{(1)}_{k|k-1}\right\rangle+\left\langle1,\underline{v}^{(0)}_{k|k-1}\right\rangle}+\right.\\
\left.\sum_{z\in
Z_k}\frac{b.\mathcal{K}(z)}{\left\langle\underline{v}^{(0)}_{k|k-1},p^{(0)}_{D}\mathcal{K}\right\rangle+\left\langle\underline{v}^{(1)}_{k|k-1},p^{(1)}_{D}g(z|\cdot)\right\rangle}\right].
\end{multline}

\section*{Acknowledgment}

The authors would like to thank Dr. William E. Hughes and
the research team from the Garvan Institute of Medical
Research, NSW, Australia, for providing the real TIRFM
sequences and their manual ground truth. The authors also thank
Dr. Anthony Dick from the University of Adelaide whose
comments and suggestions improved presentation of the paper.

\ifCLASSOPTIONcaptionsoff
  \newpage
\fi

\bibliographystyle{IEEEtran}
\bibliography{reference}

\begin{thebibliography}{10}
\providecommand{\url}[1]{#1}
\csname url@samestyle\endcsname
\providecommand{\newblock}{\relax}
\providecommand{\bibinfo}[2]{#2}
\providecommand{\BIBentrySTDinterwordspacing}{\spaceskip=0pt\relax}
\providecommand{\BIBentryALTinterwordstretchfactor}{4}
\providecommand{\BIBentryALTinterwordspacing}{\spaceskip=\fontdimen2\font plus
\BIBentryALTinterwordstretchfactor\fontdimen3\font minus
  \fontdimen4\font\relax}
\providecommand{\BIBforeignlanguage}[2]{{%
\expandafter\ifx\csname l@#1\endcsname\relax
\typeout{** WARNING: IEEEtran.bst: No hyphenation pattern has been}%
\typeout{** loaded for the language `#1'. Using the pattern for}%
\typeout{** the default language instead.}%
\else
\language=\csname l@#1\endcsname
\fi
#2}}
\providecommand{\BIBdecl}{\relax}
\BIBdecl

\bibitem{matov2011optimal}
A.~Matov, M.~M. Edvall, G.~Yang, and G.~Danuser, ``Optimal-flow minimum-cost
  correspondence assignment in particle flow tracking,'' \emph{Comput. Vis.
  Image Und.}, vol. 115, no.~4, pp. 531--540, 2011.

\bibitem{bonneau2005single}
S.~Bonneau, M.~Dahan, and L.~D. Cohen, ``Single quantum dot tracking based on
  perceptual grouping using minimal paths in a spatiotemporal volume,''
  \emph{IEEE Trans. Image Process.}, vol.~14, no.~9, pp. 1384--1395, 2005.

\bibitem{mashanov2007automatic}
G.~Mashanov and J.~Molloy, ``Automatic detection of single fluorophores in live
  cells,'' \emph{Biophys. J.}, vol.~92, no.~6, pp. 2199--2211, 2007.

\bibitem{jaqaman2008robust}
K.~Jaqaman, D.~Loerke, M.~Mettlen, H.~Kuwata, S.~Grinstein, S.~L. Schmid, and
  G.~Danuser, ``Robust single-particle tracking in live-cell time-lapse
  sequences,'' \emph{Nat. methods}, vol.~5, no.~8, pp. 695--702, 2008.

\bibitem{dewan2011tracking}
M.~Dewan, M.~Ahmad, and M.~Swamy, ``Tracking biological cells in time-lapse
  microscopy: an adaptive technique combining motion and topological
  features,'' \emph{IEEE Trans. Biomed. Eng.}, vol.~58, no.~6, pp. 1637--1647,
  2011.

\bibitem{sbalzarini2005feature}
I.~Sbalzarini and P.~Koumoutsakos, ``Feature point tracking and trajectory
  analysis for video imaging in cell biology,'' \emph{J. Struct. Biol.}, vol.
  151, no.~2, pp. 182--195, 2005.

\bibitem{padfield2011coupled}
D.~Padfield, J.~Rittscher, and B.~Roysam, ``Coupled minimum-cost flow cell
  tracking for high-throughput quantitative analysis,'' \emph{Med. Image
  Anal.}, vol.~15, no.~4, pp. 650--668, 2011.

\bibitem{meijering2006tracking}
E.~Meijering, I.~Smal, and G.~Danuser, ``Tracking in molecular bioimaging,''
  \emph{IEEE Signal Proc. Mag.}, vol.~23, no.~3, pp. 46--53, 2006.

\bibitem{house2009tracking}
D.~House, M.~Walker, Z.~Wu, J.~Wong, and M.~Betke, ``Tracking of cell
  populations to understand their spatio-temporal behavior in response to
  physical stimuli,'' in \emph{Proc. IEEE Conf. Comput. Vis. Mach. Intell.
  (CVPR 2009)}, 2009, pp. 186--193.

\bibitem{liang2011expectation}
L.~Liang, H.~Shen, P.~De~Camilli, D.~Toomre, and J.~Duncan, ``An expectation
  maximization based method for subcellular particle tracking using multi-angle
  {TIRF} microscopy,'' in \emph{Medical Imag. Comput. Computer-Assist. Interv.
  (MICCAI 2011)}, 2011, pp. 629--636.

\bibitem{nguyen2011tracking}
N.~H. Nguyen, S.~Keller, E.~Norris, T.~T. Huynh, M.~G. Clemens, and M.~C. Shin,
  ``Tracking colliding cells in vivo microscopy,'' \emph{IEEE Trans. Biomed.
  Eng.}, vol.~58, no.~8, pp. 2391--2400, 2011.

\bibitem{genovesio2006multiple}
A.~Genovesio, T.~Liedl, V.~Emiliani, W.~Parak, M.~Coppey-Moisan, and
  J.~Olivo-Marin, ``Multiple particle tracking in {3-D}+t microscopy: Method
  and application to the tracking of endocytosed quantum dots,'' \emph{IEEE
  Trans. Image Process.}, vol.~15, no.~5, pp. 1062--1070, 2006.

\bibitem{yang2012new}
L.~Yang, Z.~Qiu, A.~Greenaway, and W.~Lu, ``A new framework for particle
  detection in low-{SNR} fluorescence live-cell images and its application for
  improved particle tracking,'' \emph{IEEE Trans. Biomed. Eng.}, vol.~59,
  no.~7, pp. 2040--2050, 2012.

\bibitem{chenouard2013multiple}
N.~Chenouard, I.~Bloch, and J.-C. Olivo-Marin, ``Multiple hypothesis tracking
  for cluttered biological image sequences,'' \emph{IEEE Trans. Pattern Anal.
  Mach. Intell.}, vol.~35, no.~11, pp. 2736--2750, 2013.

\bibitem{juang2009tracking}
R.~Juang, A.~Levchenko, and P.~Burlina, ``Tracking cell motion using
  {GM-PHD},'' in \emph{Proc. IEEE Int. Symp. Biomed. Imag. (ISBI 2009)}, 2009,
  pp. 1154--1157.

\bibitem{rezatofighi2013multiple}
S.~H. Rezatofighi, S.~Gould, B.-N. Vo, K.~Mele, W.~E. Hughes, and R.~Hartley,
  ``A multiple model probability hypothesis density tracker for time-lapse cell
  microscopy sequences,'' in \emph{Inform. Process. Medical Imag. (IPMI 2013)},
  2013, pp. 110--122.

\bibitem{wood2012simplified}
T.~Wood, C.~Yates, D.~Wilkinson, and G.~Rosser, ``Simplified multitarget
  tracking using the {PHD} filter for microscopic video data,'' \emph{IEEE
  Trans. Circ. Syst. Vid.}, vol.~22, no.~5, pp. 702--713, 2012.

\bibitem{feng2011multiple}
L.~Feng, Y.~Xu, Y.~Yang, and X.~Zheng, ``Multiple dense particle tracking in
  fluorescence microscopy images based on multidimensional assignment,''
  \emph{J. Struct. Biol.}, vol. 173, no.~2, pp. 219--228, 2011.

\bibitem{rezatofighi2012application}
S.~H. Rezatofighi, S.~Gould, R.~Hartley, K.~Mele, and W.~E. Hughes,
  ``Application of the {IMM-JPDA} filter to multiple target tracking in total
  internal reflection fluorescence microscopy images,'' in \emph{Medical Imag.
  Comput. Computer-Assist. Interv. (MICCAI 2012)}, 2012, pp. 357--364.

\bibitem{li2008cell}
K.~Li, E.~D. Miller, M.~Chen, T.~Kanade, L.~E. Weiss, and P.~G. Campbell,
  ``Cell population tracking and lineage construction with spatiotemporal
  context,'' \emph{Med. Image Anal.}, vol.~12, no.~5, pp. 546--566, 2008.

\bibitem{smal2008multiple}
I.~Smal, E.~Meijering, K.~Draegestein, N.~Galjart, I.~Grigoriev, A.~Akhmanova,
  M.~Van~Royen, A.~Houtsmuller, and W.~Niessen, ``Multiple object tracking in
  molecular bioimaging by {R}ao-{B}lackwellized marginal particle filtering,''
  \emph{Med. Image Anal.}, vol.~12, no.~6, pp. 764--777, 2008.

\bibitem{smal2008particle}
I.~Smal, K.~Draegestein, N.~Galjart, W.~Niessen, and E.~Meijering, ``Particle
  filtering for multiple object tracking in dynamic fluorescence microscopy
  images: Application to microtubule growth analysis,'' \emph{IEEE Trans. Med.
  Imag.}, vol.~27, no.~6, pp. 789--804, 2008.

\bibitem{godinez2009deterministic}
W.~Godinez, M.~Lampe, S.~W{\"o}rz, B.~M{\"u}ller, R.~Eils, and K.~Rohr,
  ``Deterministic and probabilistic approaches for tracking virus particles in
  time-lapse fluorescence microscopy image sequences,'' \emph{Med. Image
  Anal.}, vol.~13, no.~2, pp. 325--342, 2009.

\bibitem{hoseinnezhad2012visual}
R.~Hoseinnezhad, B.-N. Vo, B.-T. Vo, and D.~Suter, ``Visual tracking of
  numerous targets via multi-{B}ernoulli filtering of image data,''
  \emph{Pattern Recogn.}, vol.~45, no.~10, pp. 3625--3635, 2012.

\bibitem{yuan2012object}
L.~Yuan, Y.~F. Zheng, J.~Zhu, L.~Wang, and A.~Brown, ``Object tracking with
  particle filtering in fluorescence microscopy images: Application to the
  motion of neurofilaments in axons,'' \emph{IEEE Trans. Med. Imag.}, vol.~31,
  no.~1, pp. 117--130, 2012.

\bibitem{chenouard2014objective}
N.~Chenouard, I.~Smal, F.~De~Chaumont, M.~Ma{\v{s}}ka, I.~F. Sbalzarini,
  Y.~Gong, J.~Cardinale, C.~Carthel, S.~Coraluppi, M.~Winter \emph{et~al.},
  ``Objective comparison of particle tracking methods,'' \emph{Nat. methods},
  vol.~11, pp. 281–--289, 2014.

\bibitem{Smal2010quantitative}
I.~Smal, M.~Loog, W.~Niessen, and E.~Meijering, ``Quantitative comparison of
  spot detection methods in fluorescence microscopy,'' \emph{IEEE Trans. Med.
  Imaging}, vol.~29, no.~2, pp. 282--301, 2010.

\bibitem{Burchfield2010}
J.~Burchfield, J.~Lopez, K.~Mele, P.~Vallotton, and W.~Hughes, ``Exocytotic
  vesicle behaviour assessed by {TIRFM},'' \emph{Traffic}, vol.~11, pp.
  429--439, 2010.

\bibitem{mahler2011cphd}
R.~P. Mahler, B.-T. Vo, and B.-N. Vo, ``{CPHD} filtering with unknown clutter
  rate and detection profile,'' \emph{IEEE Trans. Signal Process.}, vol.~59,
  no.~8, pp. 3497--3513, 2011.

\bibitem{beard2013multi}
M.~Beard, B.-T. Vo, and B.-N. Vo, ``Multi-target filtering with unknown clutter
  density using a bootstrap {GMCPHD} filter,'' \emph{IEEE Signal Proc. Let.},
  vol.~20, no.~4, pp. 323--326, 2013.

\bibitem{mahler2007statistical}
R.~P. Mahler, \emph{Statistical multisource-multitarget information
  fusion}.\hskip 1em plus 0.5em minus 0.4em\relax Artech House Boston, 2007,
  vol. 685.

\bibitem{bar1987tracking}
Y.~Bar-Shalom, \emph{Tracking and data association}.\hskip 1em plus 0.5em minus
  0.4em\relax Academic Press Professional, Inc., 1987.

\bibitem{blackman1986multiple}
S.~S. Blackman, \emph{Multiple-target tracking with radar applications}.\hskip
  1em plus 0.5em minus 0.4em\relax Artech House, Inc.,, 1986.

\bibitem{hue2002tracking}
C.~Hue, J.-P. Le~Cadre, and P.~P{\'e}rez, ``Tracking multiple objects with
  particle filtering,'' \emph{IEEE Trans. Aerosp. Electron. Syst.}, vol.~38,
  no.~3, pp. 791--812, 2002.

\bibitem{mahler2003multitarget}
R.~Mahler, ``Multitarget {B}ayes filtering via first-order multitarget
  moments,'' \emph{IEEE Trans. Aerosp. Electron. Syst.}, vol.~39, no.~4, pp.
  1152--1178, 2003.

\bibitem{mahler2007phd}
------, ``{PHD} filters of higher order in target number,'' \emph{IEEE Trans.
  Aerosp. Electron. Syst.}, vol.~43, no.~4, pp. 1523--1543, 2007.

\bibitem{vo2009cardinality}
B.-T. Vo, B.-N. Vo, and A.~Cantoni, ``The cardinality balanced multi-target
  multi-{B}ernoulli filter and its implementations,'' \emph{IEEE Trans. Signal
  Process.}, vol.~57, no.~2, pp. 409--423, 2009.

\bibitem{vo2010joint}
B.-N. Vo, B.-T. Vo, N.-T. Pham, and D.~Suter, ``Joint detection and estimation
  of multiple objects from image observations,'' \emph{IEEE Trans. Signal
  Process.}, vol.~58, no.~10, pp. 5129--5141, 2010.

\bibitem{vo2013labeled}
B.-T. Vo and B.-N. Vo, ``Labeled random finite sets and multi-object conjugate
  priors,'' \emph{IEEE Trans. Signal Process.}, vol.~61, no.~13, pp.
  3460--3475, 2013.

\bibitem{mahler2014statistical}
R.~P. Mahler, \emph{Advances in Statistical Multisource-Multitarget Information
  Fusion}.\hskip 1em plus 0.5em minus 0.4em\relax Artech House Boston, 2014.

\bibitem{vo2005sequential}
B.-N. Vo, S.~Singh, and A.~Doucet, ``Sequential {M}onte {C}arlo methods for
  multitarget filtering with random finite sets,'' \emph{IEEE Trans. Aerosp.
  Electron. Syst.}, vol.~41, no.~4, pp. 1224--1245, 2005.

\bibitem{vo2007analytic}
B.-T. Vo, B.-N. Vo, and A.~Cantoni, ``Analytic implementations of the
  cardinalized probability hypothesis density filter,'' \emph{IEEE Trans.
  Signal Process.}, vol.~55, no.~7, pp. 3553--3567, 2007.

\bibitem{pasha2009gaussian}
S.~Pasha, B.-N. Vo, H.~Tuan, and W.~Ma, ``A {G}aussian mixture {PHD} filter for
  jump {M}arkov system models,'' \emph{IEEE Trans. Aerosp. Electron. Syst.},
  vol.~45, no.~3, pp. 919--936, 2009.

\bibitem{rezatofighi2012new}
S.~H. Rezatofighi, R.~Hartley, and W.~E. Hughes, ``A new approach for spot
  detection in total internal reflection fluorescence microscopy,'' in
  \emph{Proc. IEEE Int. Symp. Biomed. Imag. (ISBI 2012)}, 2012, pp. 860--863.

\bibitem{vo2006gaussian}
B.-N. Vo and W.~Ma, ``The {G}aussian mixture probability hypothesis density
  filter,'' \emph{IEEE Trans. Signal Process.}, vol.~54, no.~11, pp.
  4091--4104, 2006.

\bibitem{lin2006track}
L.~Lin, Y.~Bar-Shalom, and T.~Kirubarajan, ``Track labeling and {PHD} filter
  for multitarget tracking,'' \emph{IEEE Trans. Aerosp. Electron. Syst.},
  vol.~42, no.~3, pp. 778--795, 2006.

\bibitem{panta2007novel}
K.~Panta, B.-N. Vo, and S.~Singh, ``Novel data association schemes for the
  probability hypothesis density filter,'' \emph{IEEE Trans. Aerosp. Electron.
  Syst.}, vol.~43, no.~2, pp. 556--570, 2007.

\bibitem{schuhmacher2008consistent}
D.~Schuhmacher, B.-T. Vo, and B.-N. Vo, ``A consistent metric for performance
  evaluation of multi-object filters,'' \emph{IEEE Trans. Signal Process.},
  vol.~56, no.~8, pp. 3447--3457, 2008.

\bibitem{ristic2011metric}
B.~Ristic, B.-N. Vo, D.~Clark, and B.-T. Vo, ``A metric for performance
  evaluation of multi-target tracking algorithms,'' \emph{IEEE Trans. Signal
  Process.}, vol.~59, no.~7, pp. 3452--3457, 2011.

\bibitem{rezatofighi2013framework}
S.~H. Rezatofighi, W.~T.~E. Pitkeathly, S.~Gould, R.~Hartley, K.~Mele, W.~E.
  Hughes, and J.~G. Burchfield, ``A framework for generating realistic
  synthetic sequences of total internal reflection fluorescence microscopy
  images,'' in \emph{Proc. IEEE Int. Symp. Biomed. Imag. (ISBI 2013)}, 2013.

\bibitem{rayleigh1879}
L.~Rayleigh, ``Investigations in optics, with special reference to the
  spectroscope,'' \emph{Philos. Mag.}, vol.~8, no.~49, pp. 261--274, 1879.

\bibitem{meijering2006mtrackj}
E.~Meijering, ``{MTrackJ}: A {J}ava program for manual object tracking,''
  \url{[Online]. Available:
  http://www.imagescience.org/meijering/software/mtrackj/}.

\end{thebibliography}

\end{document}